\documentclass{article}
\usepackage{graphicx} % Required for inserting images
\usepackage{fullpage}
\usepackage[numbers, sort&compress]{natbib}
\usepackage[hyphens]{url}
\usepackage{hyperref}
\usepackage{authblk}
\usepackage[dvipsnames]{xcolor}
\usepackage{rotating}   % for sidewaysfigure
\usepackage{subcaption} % for subfigure support
\usepackage{caption}    % for \Continued
\usepackage{algorithm}
\usepackage{algorithmic}
\usepackage{booktabs}
\usepackage[export]{adjustbox}
\usepackage{array} 
\usepackage{wrapfig}

\usepackage{amsmath}
\usepackage{amssymb}
\usepackage{mathtools}
\usepackage{amsthm}
\theoremstyle{plain}
\newtheorem{theorem}{Theorem}[section]
\newtheorem{proposition}[theorem]{Proposition}
\newtheorem{lemma}[theorem]{Lemma}
\newtheorem{corollary}[theorem]{Corollary}
\theoremstyle{definition}
\newtheorem{definition}[theorem]{Definition}

\theoremstyle{remark}

\title{Characterizing Predictive Multiplicity in Ranking and Its Fairness Implications}
\author[]{Ainhize Barrainkua} 
\author[]{Mikel Malagón} 

\affil[]{University of the Basque Country (EHU), \texttt{\{ainhize.barrainkua,mikel.malagon\}@ehu.eus}}

\date{}

\begin{document}

\maketitle

\newcommand{\topk}{{\boldsymbol{\pi}_{1:K}}}

\newcommand{\settopk}{\mathcal{C}_{\topk}}

\newcommand{\piref}{{\pi^\bigstar}}

\newcommand{\setref}{\mathcal{C}^\bigstar_{K}}

\begin{abstract}
Prediction problems often admit competing models that achieve comparable performance yet produce conflicting outputs, a phenomenon known as predictive multiplicity. While this challenge has been extensively studied in classification tasks, its implications remain largely unexplored in ranking frameworks, which are critical for high-stakes societal decisions such as the allocation of welfare resources and social assistance.
This paper investigates the phenomenon of predictive multiplicity in score-based ranking problems. We introduce formal definitions to capture multiplicity in ranking settings, supported by foundational theoretical guarantees. Furthermore, we examine the dual relationship between predictive multiplicity and algorithmic fairness in ranking problems: we analyze how model multiplicity can act as a potential source of unfairness, while demonstrating how exploring this decision space provides a principled avenue for discovering less discriminatory rankings. Finally, we complement our theoretical contributions with insightful empirical evaluations on both synthetic and real-world datasets, validating our analytical findings and illustrating their practical implications.
Taken together, this work highlights the real-world relevance of predictive multiplicity in ranking 
and sheds light on the potential of leveraging this effect to achieve fairer outcomes.
\end{abstract}

\section{Introduction}

In the era of algorithmic decision-making, predictive models are increasingly deployed to arbitrate high-stakes societal decisions, ranging from criminal justice and healthcare to the allocation of welfare resources and social assistance. A foundational assumption underlying this paradigm is that optimization—finding a model that maximizes predictive performance—yields a singular, objective standard for decision-making. However, a growing body of empirical and theoretical work challenges this premise \citep{marx2020predictive,hsu2022rashomon, watson2023predictive, behzad2025reconciling}, revealing that prediction problems frequently admit competing models that achieve comparable performance while generating conflicting outputs, a phenomenon known as \textit{predictive multiplicity} \citep{marx2020predictive}.

Rooted in Leo Breiman's seminal observation of the "Rashomon Effect" \citep{breiman2001statistical}, predictive multiplicity occurs when a dataset supports a vast multiplicity of models with comparable predictive accuracy or empirical risk, yet these models disagree substantially on individual instances. While the Rashomon effect has traditionally been studied through the lens of classical model interpretability and feature selection, its implications have profound real-world consequences when models are operationalized in downstream decision pipelines.

Although this phenomenon has been extensively studied in standard classification tasks \citep{marx2020predictive}—where inconsistent model predictions introduce instability and arbitrariness into binary or multiclass outcomes—its implications remain largely unexplored in ranking frameworks. Ranking tasks present a fundamentally different mathematical structure: rather than merely categorizing instances, ranking systems sort individuals or items into a prioritized sequence. In resource-allocation settings, such as the distribution of social assistance or welfare benefits, a ranking does not just dictate a label; it establishes a strict hierarchy of need and eligibility under conditions of scarcity \cite{zehlike2022fairness}. Consequently, when multiple rankings achieve near-optimal utility yet drastically reorder candidates, the resulting multiplicity directly dictates who receives vital support and who is systematically overlooked.
This vulnerability is amplified across a wide array of high-stakes domains—including healthcare prioritization for scarce medical resources, hiring and recruitment under candidate caps, credit and loan approvals, university admissions, visa processing and social welfare or public housing allocation. 
%In each of these settings, when multiple models achieve near-optimal performance yet drastically reorder candidates, the resulting instability means that an individual's access to vital life outcomes depends more on arbitrary algorithmic choices than on objective need or qualification.

Motivated by these severe social implications, this work establishes a formal framework for studying predictive multiplicity in score-based ranking problems. Specifically, we introduce formal definitions to capture multiplicity in general ranking settings, backed by foundational theoretical results that characterize the structure and behavior of the Rashomon set in ordered spaces. Furthermore, we examine the intricate, dual relationship between predictive multiplicity and algorithmic fairness: first, as a potential source of unfairness—where arbitrary selection among equally valid rankings drastically alters individual outcomes—and second, as an avenue for bias mitigation. By systematically searching across utility-equivalent rankings that yield different candidate orderings, we demonstrate how predictive multiplicity can be leveraged to discover rankings that maximize fairness guarantees.

To complement our theoretical contributions, we conduct comprehensive empirical evaluations across both synthetic benchmark scenarios and real-world datasets, validating our theoretical results and illustrating the practical impact of multiplicity in real-world decision-making.\footnote{Code and data are accessible at: \url{https://anonymous.4open.science/r/fair-rankings-C2C3}.}
Taken together, our theoretical insights and empirical evidence underscore the urgent need to account for predictive multiplicity in the design, evaluation, and auditing of automated ranking systems deployed in society.

\textbf{Our contributions.} 
We begin by establishing the theoretical background (Section~\ref{sec:theoretical-background}) and situating our study within the related literature (Section~\ref{sec:related-works}). We then formalize the mathematical framework for predictive multiplicity in score-based ranking problems by defining the Rashomon set (Section~\ref{sec:rashomon-set}), proving theoretical bounds on its size to quantify the degree of multiplicity (Section~\ref{sec:rashomon-set-size}), and characterizing the decision boundary that governs resource access (Section~\ref{sec:decision-boundary}). Next, we analyze the dual impact of multiplicity on ranking fairness—evaluating it both as a source of individual prediction instability (Section~\ref{sec:rashomon-flip-prob}) and as a search space for discovering less discriminatory rankings (Section~\ref{sec:rashomon-group-fairness}). We complement these analytical findings with empirical evaluations on both synthetic toy scenarios and real-world datasets (Sections~\ref{sec:rashomon-empirical-insights} and \ref{sec:experiments}). Finally, we summarize our main findings and outline future research directions (Section~\ref{sec:conclusions}).

\section{Theoretical Background}
\label{sec:theoretical-background}

\paragraph{Notation.} Sets are denoted by calligraphic letters ($\mathcal{X}$), random variables by uppercase letters ($X$), and instantiations by lowercase letters ($x$). Bold lowercase and uppercase letters represent vectors and matrices, respectively (e.g., $\boldsymbol{v}$ and $\boldsymbol{M}$).

We consider  a set of $N$ candidates $\mathcal{C} = \{(\mathbf{x}_i, s_i, y_i) \}_{i=1}^N$, where each candidate is represented by the triplet $(\mathbf{x}, s, y)$. Here, $\mathbf{x} \in \mathcal{X}  \subseteq \mathbb{R}^d$ denotes the vector of non-sensitive attributes, $s \in \mathcal{S} = \{1, \ldots, A\}$ represents the sensitive group membership, and $y \in \mathcal{Y}$ denotes a relevance, utility, or qualification score associated with the candidate.
We assume that the input data corresponds to individuals, e.g., persons seeking educational, employment, or financial opportunities, or being prioritized for access to goods and services.
The sensitive groups may be overlapping and defined by one or multiple sensitive attributes (e.g., gender and race) and can therefore capture intersectional group identities.

Formally, we denote a ranking $\boldsymbol{\pi}$ as a permutation over the set of candidates $\mathcal{C}$. A permutation is a bijection of a set onto itself, i.e., $\boldsymbol{\pi} \; : \; \mathcal{C} \rightarrow \mathcal{C}$. It can be interpreted as a particular ordering of the members of a set. The set of all permutations of a set with $N$ elements forms the symetrics group $\mathbb{S}_N$.  
The resulting ranking induces an ordered list in which higher-ranked individuals receive greater visibility, priority, or access to opportunities. We denote by $\boldsymbol{\pi} = ( \pi_1, ..., \pi_N )$, a ranking that places candidate $\pi_i$ at rank $i$, where $\boldsymbol{\pi}(i) = \pi_i$, and $\boldsymbol{\pi}^{-1}(w)$ denotes the rank of candidate $w$ in $\boldsymbol{\pi}$. In many applications, we are interested in a sub-ranking of $\boldsymbol{\pi}$ that contains its best-ranked $K$ elements, where $K \leq N$,
%. This sub-ranking is called the top-$K$ and 
denoted $\topk$.  
In this work, we will focus on the context where a top-$K$ must be selected: 
\begin{equation}
    \settopk = \{\topk(i): i\in[1, K]\}.
\end{equation}
In these contexts, the ranking effectively determines allocation decisions, such that small differences in ranking positions may lead to substantially different outcomes. Consequently, disparities in ranking performance across sensitive groups can translate directly into unequal access to limited opportunities and resources.

Although the analysis presented in this work can be extended to entities other than people, the fairness considerations and guarantees we study are primarily relevant in settings where the data describes individuals. For consistency, we therefore refer to the set of individuals considered in a ranking task as candidates.

In particular, this work focuses on the Rashomon effect in \textit{score-based} ranking problems. Given the set of candidates $\mathcal{C}$, 
%the goal of learning to rank is to learn a scoring function that assigns a score to each candidate and thereby induces an ordering over $\mathcal{C}$. Let $f:\mathcal{X}\rightarrow\mathbb{R}$ denote a scoring function from a hypothesis class $\mathcal{F}$. The score assigned to candidate $i$ is $\beta_i=f(\mathbf{x}_i)$,
each candidate $i$ is assigned a score $\beta_i$,
and the resulting ranking $\boldsymbol{\pi}_f$ is obtained by sorting the candidates according to their scores, such that 
% $\beta_{\boldsymbol{\pi}_f(1)}\geq \beta_{\boldsymbol{\pi}_f(2)}\geq\cdots\geq \beta_{\boldsymbol{\pi}_f(N)}.$
\begin{equation}
\beta_{\boldsymbol{\pi}_f(1)}\geq \beta_{\boldsymbol{\pi}_f(2)}\geq\cdots\geq \beta_{\boldsymbol{\pi}_f(N)}.
\end{equation}
Under this setting, the \textit{learning-to-rank} paradigm, aims to learn a scoring function $f:\mathcal{X}\rightarrow\mathbb{R}$ from a hypothesis class $\mathcal{F}$  that assigns a score to each candidate, and thereby induces an ordering over $\mathcal{C}$. In this work, however, we do not focus on how these scores, and the resulting rankings, are obtained. Instead, we consider the candidates' true utility scores, $y_i$, and set the focus on the ranking level. Specifically, we investigate whether multiple top-$K$ configurations can achieve equal or similar utility while comprising different sets of candidates. Our aim is to formally characterize this phenomenon and examine its implications at the candidate level.

\textit{But how can we evaluate the "goodness" of a ranking?}
Assuming that the score $y$ encodes the appropriateness of the candidates, the utility $u$ of a top-k set $\settopk$ can be defined as the sum of scores of its elements \cite{zehlike2022fairness}:
\begin{equation}
    u(\settopk) = \sum_{\pi\in\settopk} y_{\boldsymbol{\pi}}.
    \label{eq:ranking_utility}
\end{equation}

\section{Related Works}
\label{sec:related-works}

This work lies at the intersection of predictive multiplicity, algorithmic fairness, and fair ranking. While predictive multiplicity and fairness have been extensively studied in classification, their intersection in ranking remains largely unexplored. In particular, little is known about how the existence of different rankings with similar utility affect the fairness of ranked outcomes.

Predictive multiplicity, also known as the \textit{Rashomon effect}, refers to the existence of multiple models with comparable predictive performance but potentially conflicting predictions or explanations \cite{marx2020predictive, watson2023predictive, behzad2025reconciling, ganesh2025systemizing, haghighat2026resolving}. This raises the question of how to select among equally accurate models, particularly in high-stakes applications where conflicting predictions may affect individuals and demographic groups \cite{binns2018s}. However, multiplicity also offers an opportunity to prioritize fairness, robustness, or interpretability without sacrificing predictive performance \cite{black2022model}. While prior work has primarily focused on classification tasks, many real-world applications involve ranking individuals for scarce resources, such as social housing or financial assistance \cite{zehlike2022fairness}. In these settings, different orderings may produce rankings with similar utility, making the fairness implications of ranking multiplicity essential to understand.

Fairness in ranking has been assessed through metrics that capture different dimensions of ranked outcomes, including representation  \cite{yang2017measuring, celis_et_al:LIPIcs:2018:9032, stoyanovich2018online}, utility \cite{yang_ijcai2019p836}, exposure \cite{Singh2018fairness, singh2019policy, Sapiezynski2019quantifying, raj2022measuring}, pairwise \cite{beutel2019fairness, narasimhan2020pairwise}, and counterfactual fairness \cite{yang_et_al:LIPIcs.FORC.2021.7}. These metrics address disparities arising from pre-existing, technical, and emergent biases.
%, including unequal representation and exposure, relative ranking positions, and the effects of sensitive attributes. 
For a comprehensive review, see \citet{schumacher2025properties}. Correspondingly, fairness-enhancing interventions operate at different stages of the ranking pipeline: score distributions, through score correction and counterfactual reasoning \cite{kleinberg_et_al:LIPIcs:2018:8323, celis2020interventions, yang_et_al:LIPIcs.FORC.2021.7}; learned scoring functions, by finding fair functions close to the originally learned \cite{asudeh2019designing}; and ranking outputs, by imposing representation or utility constraints on top-$K$ results or ranking prefixes \cite{yang2017measuring, celis_et_al:LIPIcs:2018:9032, yang_ijcai2019p836}. Despite these advances, existing methods generally seek a single fair scoring function or ranking, overlooking the possibility that multiple similarly performing rankings may induce different outcomes at individual-level. We address this gap by characterizing predictive multiplicity in ranking and investigating its implications regarding algorithmic fairness. 

A more detailed description on the related works can be found in Appendix \ref{ap:related-works-long}.

\section{Characterizing the Rashomon Effect in Rankings}
\label{sec:rashomon_ranking}

As a first step toward characterizing the Rashomon effect in score-based ranking problems, we focus on the space of possible top-$K$ selections induced by different rankings, without imposing any restrictions on the family of scoring functions used to generate those rankings. This allows us to study the phenomenon at the level of the ranking itself, independently of the particular modeling choices used to produce it. 
%In practice, however, rankings are generated by selecting a scoring function from a specified hypothesis class or model family, and this restriction can substantially constrain the set of attainable rankings and, consequently, the structure of the Rashomon set. We therefore first develop a model-agnostic characterization of the Rashomon effect at the level of top-$K$ selections. In subsequent sections, we incorporate the constraints imposed by specific scoring-function families and study how the choice of model class affects the resulting Rashomon set.

%In this section we introduce the main definitions and theoretical results regarding the Rashomon effect and predictive multiplicity for ranking-based problems. 
% In the case of classification problems, the definition of the Rashomon set has been provided through two different perspectives. In all the cases, a reference model was considered: either an already operating/optimized model or the optimal classifier (i.e., the Bayes classifier). Then, they defined the Rashomon set as the set of models/classifiers that had a error difference of $\epsilon$ from that reference classifier. 
The Rashomon effect is characterized with respect to a reference ranking
%model (or ranking in this case), 
denoted by $\piref$. In the case where $\piref$ is an optimal ordering of the set of candidates $\mathcal{C}$, $\piref$ has been also referred to as the Bayes optimal ranking \cite{cossock2008statistical}, and satisfies the following condition on the true utilities $y$ of the candidates:
% $y_{\piref(1)} \geq ... \geq y_{\piref(N)}$.
\begin{equation} \label{eq:ordered-scores}
    y_{\piref(1)} \geq ... \geq y_{\piref(N)}.
\end{equation}

To investigate the Rashomon effect in ranking problems independently of the underlying scoring function, we first formalize the Rashomon set for ranking-based tasks. We then derive theoretical results characterizing its properties, including its size, which serves as a proxy for the magnitude of the Rashomon effect. We further characterize the decision threshold and complement our theoretical analysis with experiments on a toy problem to illustrate and validate our findings.

% \color{orange}
% We have to further formalize this: The rashomon set has to count all the rankings with similar utility, and Predictive multiplicity the (degree of) disagreement of these rankings with similar utility. Say that in ranking based problems they are equal??  
% \color{black}

%For that, we first define the Bayes-optimal ranking, which is they employed to define the Rashomon set, and analyze its size (and other characteristics?). 
% We then introduce different metrics to quantify the multiplicity, and analyze the problem of the practical search of the Rashomon set.

\subsection{The Rashomon Set}
\label{sec:rashomon-set}

%As specified in the introduction, w
We consider the setting where we are given a ranking composed of a set of $N$ candidates, and
% $\mathcal{C}$, a scoring function assigns a utility score to each candidate, thereby inducing a ranking. 
the top-$K$ candidates are then selected according to this ranking. When only the membership of the selected set matters and the ordering among the selected candidates is ignored, each possible selection corresponds to a combination of $K$ candidates from $N$. Hence, the total number of possible top-$K$ selections is $\binom{N}{K} = \frac{N!}{K!(N-K)!}.$
% \begin{equation}
% \binom{N}{K} = \frac{N!}{K!(N-K)!}.
% \end{equation}
Among all those possible rankings, the $\epsilon$-level (model-agnostic) Rashomon set consists solely of those rankings whose utility is at most $\epsilon$ lower than the Bayes-optimal ranking, as formally defined below.

%the Rashomon set only comprises those whose utility is at most $\epsilon$ worse than that of the Bayes-optimal ranking. A formal definition is introduced below. 

\begin{definition} [$\epsilon$-Level Model-Agnostic Rashomon Set for top-$K$]
    Let $\piref$ be the Bayes optimal permutation where the instances are sorted according to their true utility (i.e., $y_{\piref(1)} \geq ... \geq y_{\piref(N)}$) where $\setref$ is the set of its top-K candidates according to $\piref$,
    $\mathcal{C}_K$ be any set top-K set, 
    %(i.e., the output from a specific ranking algorithm), 
    and the utility of a ranking be defined as in Equation~\eqref{eq:ranking_utility}.
    %For a permutation $\pi$, let $\pi(i)$ denote the instance placed at position $i$. 
    Then, the $\epsilon$-level Rashomon set for Ranking is defined as:
    \begin{equation}
        \mathcal{R}_{\epsilon}^K = \left\{ \mathcal{C}_K \subset \mathcal{C} \; : \; |\mathcal{C}_K| = K, \left( \sum_{y \in \setref \setminus \mathcal{C}_K}y \; - \sum_{y' \in \mathcal{C}_K \setminus \setref} y' \right) \leq \epsilon \right\}
        \label{eq:rashomon_set}
    \end{equation}
\end{definition}

In the problem setting under consideration, the presence of the Rashomon effect inevitably leads to predictive multiplicity, reflected as individual-level disagreements among models with similar utility. Generating an alternative top-$K$ ranking requires exchanging instances, which guarantees discrepancies at the individual level. Ultimately, the magnitude of this multiplicity is determined by the size of the Rashomon set.

\subsection{The Size of the Rashomon Set}
\label{sec:rashomon-set-size}

Let us assume that $m$ items are replaced across across $\setref$ (the Bayes-optimal top-$K$ set) and $\bar{\setref}$ (the complementary of Bayes-optimal top-$K$ set).
%, there are $\binom{K}{m}\binom{N-K}{m}$ different such sets. 
Therefore, an exact characterization of the cardinality of the Rashomon set is:
\begin{equation}
    |\mathcal{R}_{\epsilon}^K| = \sum_{m=0}^K \sum_{\substack{\mathcal{S} \subseteq \setref \\ |\mathcal{S}| = m }} \sum_{\substack{\mathcal{T} \subseteq [N] \setminus \setref \\ |\mathcal{T}| = m }} \boldsymbol{1} \bigg ( \sum_{i \in \mathcal{S}}y_{i} \; - \sum_{i \in \mathcal{T}} y_{j} \leq \epsilon \bigg )
\end{equation}
where $[N]=\{1, 2, ..., N \}$. Therefore, the size of the Rashomon set will depend on the number of total candidates, the value of selected top candidates $K$, the tolerance $\epsilon$ and how the utilitiy values $y$ are distributed across the population. 
Importantly, the size of the Rashomon set is governed by how sharply utilities decay around the top-$K$ boundary, which is formalized in the following theorem. 

\begin{theorem}[Upper bound on the size of the $\epsilon$-level Model-Agnostic Rashomon set]
Let $\setref$ denote the Bayes-optimal top-$K$ set obtained from the Bayes-optimal permutation $\piref$ induced by utilities
$y_{\piref(1)} \geq ... \geq y_{\piref(N)}$. Let us define the utility gap at the boundary as $\gamma_K = y_K - y_{K+1}$. Let $\mathcal{R}_\epsilon^K$ denote the $\epsilon$-level Rashomon set of top-$K$ selections defined by Equation~\eqref{eq:rashomon_set}. 
Then the cardinality of the Rashomon set satisfies
\begin{equation}
    |\mathcal{R}_\epsilon^K|
\le
\sum_{m=0}^{ M }
\binom{K}{m}\binom{N-K}{m},
\label{eq:rashomon_size_bound}
\end{equation}
where $M = \min \{ \lfloor \epsilon / \gamma_K \rfloor, K, N-K \} $ is the maximum number of distinct elements between two top-$K$ sets with regret tolerance $\epsilon$.
% \[
% |\mathcal{R}_\epsilon|
% \le
% \sum_{m=0}^{\lfloor \epsilon / \gamma_K \rfloor}
% \binom{K}{m}\binom{N-K}{m}.
% \]
\label{th:rashomon_set_size}
\end{theorem}

For the proof of Theorem \ref{th:rashomon_set_size} refer to Appendix \ref{ap:proof_rashomon_size}.  
It is worth noting that this formulation provides an upper bound rather than the exact cardinality of the Rashomon set, because the swap capacity $M$ relies on a uniform worst-case cost assumption governed by the boundary gap $\gamma_K$. In practice, utility gaps between non-adjacent items are typically larger, meaning that many $m$-swap configurations incur an actual regret exceeding $\epsilon$ and are therefore overcounted in the combinatorial sum. Nevertheless, this relaxation yields a tractable, distribution-agnostic bound that captures the fundamental geometric limits of ranking instability without requiring explicit knowledge of the underlying hypothesis class.
Furthermore, the following proposition formalizes the effect of the different parameters on the size of the Rashomon set.

\begin{proposition}[Dependence of the Rashomon set size on problem parameters]
Let $\mathcal{R}_\epsilon^K$ denote the $\epsilon$-level model-agnostic Rashomon set of top-$K$ selections, and let
\begin{equation}
|\mathcal{R}_\epsilon^K|
\le
\sum_{m=0}^{M}
\binom{K}{m}\binom{N-K}{m},
\qquad
M = \min \left( \left\lfloor \frac{\epsilon}{\gamma_K} \right\rfloor, K, N-K\right),
\quad
\gamma_K = y_K - y_{K+1}.
\end{equation}
Then the following holds:

\begin{enumerate}
    \item For fixed $K$ and $M$, the Rashomon set size scales polynomially in $N$ as $|\mathcal{R}_\epsilon^K| = O(N^M)$.
    % \begin{equation*}
    %     |\mathcal{R}_\epsilon^K| = O(N^M).
    % \end{equation*}

    \item In the dense regime $K = \alpha N$ with $\alpha \in (0,1)$ and fixed $M$, $|\mathcal{R}_\epsilon^K| = O\!\left(N^{2M}\right),$
    % \begin{equation*}
    %     |\mathcal{R}_\epsilon^K| = O\!\left(N^{2M}\right),
    % \end{equation*}
    with leading-order dependence proportional to $\alpha^M (1-\alpha)^M$.

    \item If $\gamma_K = 0$ for $\forall K \leq N$, i.e. in case the utilities are degenerate, then $|\mathcal{R}_\epsilon^K| = \binom{N}{K}.$
    % \begin{equation*}
    %     |\mathcal{R}_\epsilon^K| = \binom{N}{K}.
    % \end{equation*}

    \item If $\epsilon \to 0$, then $M=0$ and $|\mathcal{R}_\epsilon^K| = 1$.
\end{enumerate}
\label{prop:rashomon_size_param}
\end{proposition}

The proof of the proposition can be found in Appendix~\ref{ap:proof_param_size}. Note that, according to these theoretical results, in cases where $K$ is not fixed but rather scales linearly with $N$ (i.e., $K = \alpha N$), both sides of the boundary (the top-$K$ and its complementary) grow with the problem size. Consequently, the Rashomon set grows faster, at rate roughly $O(N^{2M})$, reflecting the increased number of possible exchanges between selected and non-selected items.

Besides, another key implication of the above analysis is that, as previously stated, the structure of the Rashomon set is highly sensitive to the utility gap at the top-$K$ boundary, i.e., $\gamma_K = y_K - y_{K+1}$. In the extreme case where all utilities are identical, i.e., $y_1 = \cdots = y_N,$
every subset of size $K$ attains the same utility, and consequently all top-$K$ selections belong to the Rashomon set, yielding $|\mathcal{R}_\epsilon^K| = \binom{N}{K}, \forall \epsilon \ge 0$.
More generally, a similar phenomenon arises whenever the utility landscape is locally flat around the decision boundary. In particular, when the utilities of candidates near ranks $K$ and $K+1$ are close, i.e., $\gamma_K \approx 0,$
exchanges of individuals across the top-$K$ boundary induce only negligible changes in aggregate utility. As a result, many substantially different rankings (or selected subsets) achieve nearly identical objective value, even under small $\epsilon$-perturbations of the Bayes-optimal solution.
While these alternative rankings are almost indistinguishable from an aggregate utility perspective, they may correspond to substantially different outcomes at the individual level: a candidate may either be selected (top-$K$) or rejected (outside top-$K$), despite there being no meaningful difference in utility between the two cases. This creates a form of selection instability or threshold unfairness, where small perturbations in the ranking function or data can lead to large changes in individual outcomes while producing only negligible changes in global utility. This effect will be further analyzed in Section~\ref{sec:fairness-analysis}.

% EXAMPLE GOES HERE OR ON A SEPARATE SECTION?

% \begin{example}
%     To develop intuition for how the utility distribution influences the size of the Rashomon set, we consider a toy example with 5 candidates ($N=5$) and a top-$K$ set of size 3 ($K=3$). Since only membership in the top-$K$ set matters, and the exact ordering of candidates is ignored, the number of distinct top-$K$ configurations is given by $\binom{5}{3}=\frac{5!}{3!2!}=10$. Consequently, the Rashomon set can contain at most 10 distinct top-$3$ configurations; that is, its maximum possible size is 10. Moreover, let us assume that candidate utilities are drawn from a normal distribution with mean zero while varying the variance, i.e., $y \sim \mathcal{N}(0, \sigma^2)$. By varying the variance of the utility distribution, we investigate how the spread of utility values affects the size of the Rashomon set. When the variance is small, the sampled utilities are concentrated around the mean, resulting in similar utility values across candidates. As the variance increases, the utilities become more dispersed, producing larger differences between candidates' utility values. This is depicted in Figure ...  \color{red} this explanation is not finished \color{black}
% \end{example}

\subsection{The Decision Boundary Gap}
\label{sec:decision-boundary}

As discussed in the previous section and established by Theorem \ref{th:rashomon_set_size}, the size of the Rashomon set depends on the decision boundary gap ($\gamma_K$) of the Bayes optimal ranking, defined as the score difference between the lowest-scoring candidate in the top-$K$ and the highest-scoring candidate excluded from the top-$K$ in such ranking. The following theorem formally characterizes the expectation of the decision boundary gap and that of the upper bound on the Rashomon-set size for the case when the utilities are i.i.d. from a continuous distribution.

\begin{theorem}[Expected boundary gap for top-$K$ rankings]
    
Let $Y_1,\ldots,Y_N$ be i.i.d. random variables drawn from a continuous distribution with cumulative distribution function $F_Y$ and probability density function $f_Y$. Let $Y_{[1]}\ge Y_{[2]}\ge\cdots\ge Y_{[N]}$ denote their descending order statistics. For $K\in\{1,\ldots,N-1\}$, we define the boundary gap between the $K$-th and $(K+1)$-th ranked items as $\gamma_K=Y_{[K]}-Y_{[K+1]}.$
Assuming the expectation exists, the expected boundary gap is
\begin{equation}
\overline{\gamma}_K = \mathbb{E}[\gamma_K] =N\binom{N-1}{K}
\int_{-\infty}^{\infty}
F_Y(y)^{N-K-1}f_Y(y)
\left(
\int_y^\infty
\bigl(1-F_Y(t)\bigr)^K\,dt
\right)dy.
\label{eq:expected-gap-continuous}
\end{equation}
\label{th:boundary-gap-continuous}
\end{theorem}

\begin{corollary}[Upper bound on the expected Rashomon-set size]
Let $\gamma_K>0$ almost surely denote the boundary gap between the $K$-th and $(K+1)$-th order statistics, and let $F_{\gamma_K}$ be its cumulative distribution function. Under the assumptions of Theorem \ref{th:boundary-gap-continuous}, the expected size of the top-$K$ Rashomon set satisfies
\begin{equation}
\mathbb E\!\left[|\mathcal R_\epsilon^K|\right]
\leq
1+\sum_{m=1}^{\min(K,N-K)}
\binom Km\binom{N-K}{m}
F_{\gamma_K}\!\left(\frac{\epsilon}{m}\right).
\label{eq:upper-bound-rashomon-continuous}
\end{equation}
\label{cor:corollary-up-size-continuous}
\end{corollary}

The identity from Theorem \ref{th:boundary-gap-continuous} is derived from the standard joint-density formula for order statistics \cite{casella2024statistical} (Theorem 5.4.6). The  proofs of Theorem \ref{th:boundary-gap-continuous} and Corollary \ref{cor:corollary-up-size-continuous} can be found in Appendix \ref{ap:boundary-gap-continuous} and Appendix \ref{ap:corollary-up-size-continuous}, respectively. 
%Besides, we can further define the upper bound on the size of the Rashomon set. 
%The proof of the corollary is provided in Appendix \ref{ap:corollary-up-size-continuous}. 
Note that, in practice, the underlying distribution from which the utilities are sampled is generally unknown. Consequently, its probability density function and cumulative distribution function cannot be specified, making it infeasible to evaluate Equations \ref{eq:expected-gap-continuous} and \ref{eq:upper-bound-rashomon-continuous} directly. As a first step toward obtaining interpretable results, we therefore investigate the special case in which the utilities are assumed to follow a Gaussian distribution. In particular, we derive an approximation of the expected boundary gap in Theorem \ref{th:boundary-gap} and an asymptotic approximation to the upper bound on the expected Rashomon-set size in Corollary \ref{cor:rashomon-size-approx-gaussian}.

\begin{theorem} [Approximate expected boundary gap for Gaussian utilities] 
Let $Y_1,\ldots,Y_N$ be i.i.d. random variables drawn from $\mathcal N(\mu,\sigma^2)$, with $\sigma>0$, and let $Y_{[1]}\ge\cdots\ge Y_{[N]}$ denote their descending order statistics. For $K\in\{1,\ldots,N-1\}$, define the boundary gap as $\gamma_K=Y_{[K]}-Y_{[K+1]}$. Then, using Blom's approximation for the expected normal order statistics \citep{blom1958statistical}, the expected boundary gap can be approximated by

\begin{equation} 
\overline{\gamma}_K = \mathbb E[\gamma_K] \approx \sigma\Bigg[ \Phi^{-1}\left( \frac{N-K+\frac{5}{8}}{N+\frac{1}{4}} \right) \Phi^{-1}\left( \frac{N-K-\frac{3}{8}}{N+\frac{1}{4}} \right) \Bigg]
\end{equation}
where $\Phi$ denotes the CDF of the standard normal distribution, , i.e., $\Phi(x) = P(Z \leq x), \; \; Z \sim \mathcal{N}(0,1).$
\label{th:boundary-gap}
\end{theorem}

\begin{corollary}[Approximate upper bound on the expected Rashomon-set size] \label{coro:set-size-normal}
Suppose $Y_1,\ldots,Y_N$ are independent and identically distributed as
$\mathcal{N}(\mu,\sigma^2)$, with $\sigma>0$. Let
$\gamma_K=Y_{[K]}-Y_{[K+1]}$ denote the boundary gap, and let
$\mathcal R_\epsilon^K$ denote the top-$K$ Rashomon set. For sufficiently
large $N$ 
%($N\to\infty$), 
and $K/N\to\rho\in(0,1)$, an
asymptotic approximation to the expected upper bound on the Rashomon-set
size is
\begin{equation}
\mathbb E\!\left[|\mathcal R_\epsilon^K|\right] \lesssim\ 1 +\sum_{m=1}^{\min(K,N-K)} \binom Km\binom{N-K}{m} \left[ 1-\exp\left( -\frac{N\epsilon}{m\sigma} \phi\left(\Phi^{-1}\left(1-\frac{K}{N}\right) \right) \right) \right].
\label{eq:approx-gaussian-rashomon-bound}
\end{equation}
Here, $\Phi$ and $\phi$ denote the standard Gaussian CDF and density,
respectively. 
% The symbol $\lesssim$ emphasizes that the expression is an asymptotic approximation to the bound, rather than a rigorous finite-sample upper bound.
\label{cor:rashomon-size-approx-gaussian}
\end{corollary}

% The following theorem characterizes this gap in the Gaussian setting. 

% \begin{theorem} [Expected boundary gap for Gaussian utilities]
%     Let $y_{\piref(1)} \geq ... \geq y_{\piref(N)}$ the utilities ordered according to the Bayes-optimal permutation. Under the assumption that $y_i \overset{\text{iid}}{\sim} \mathcal{N}(\mu, \sigma^2)$, the expected decision boundary gap can be characterized as
%     \begin{equation}
%         \overline{\gamma}_K = \mathbb{E}[\gamma_K] \approx \sigma \bigg [ \Phi^{-1} \bigg ( \frac{N-K + \frac{5}{8}}{N + \frac{1}{4}} \bigg ) - \bigg ( \frac{N-K - \frac{3}{8}}{N + \frac{1}{4}} \bigg ) \bigg ] 
%     \end{equation}
%     where $\Phi$ is the cumulative distribution function (CDF) of the standard normal distribution, i.e., $\Phi(x) = P(Z \leq x), \; \; Z \sim \mathcal{N}(0,1).$
%     \label{th:boundary-gap}
% \end{theorem}

The proofs of Theorem~\ref{th:boundary-gap} and Corollary~\ref{cor:rashomon-size-approx-gaussian} can be found in Appendix~\ref{ap:proof_boundary_gap} and Appedix~\ref{ap:rashomon-size-approx-gaussian}, respectively. 
Note that, according to the results of Theorem~\ref{th:boundary-gap}, the mean $\mu$ cancels out, and the expected decision boundary gap depends only on the standard deviation $\sigma$. This is consistent with the intuition that adding a constant to all utilities neither alters their relative ordering nor the pairwise utility gaps. Equivalently, the expected boundary gap can be expressed as, $\overline{\gamma}_K = \mathbb{E}[\gamma_K] \approx \sigma C_{N,K}$, 
% \begin{equation}
%     \mathbb{E}[\gamma_K] \approx \sigma \times C_{N,K}
% \end{equation}
where $C_{N,K}$ is a constant that depends only on $N$ and $K$.

Besides, Corollary~\ref{cor:rashomon-size-approx-gaussian} highlights the influence of the tolerance $\epsilon$, the utility variance $\sigma^2$, and the position of the top-$K$ boundary on the potential multiplicity of near-optimal rankings. For a fixed utility scale, the approximate upper bound is nondecreasing in $\epsilon$, reflecting the fact that larger tolerances allow more alternative top-$K$ sets to qualify as near-optimal. Conversely, increasing $\sigma$ while keeping $\epsilon$ fixed decreases the bound, as greater dispersion in the utilities makes small boundary gaps less likely under the adopted spacing approximation. The dependence on $K$ is captured by both the combinatorial coefficients, which count the possible exchanges across the top-$K$ boundary, and the Gaussian density evaluated at the corresponding quantile. 
%Moreover, the expression is symmetric under the exchange of $K$ and $N-K$. 
% Lastly, 
% %for sufficiently small tolerances, a first-order expansion of the exponential term shows that the excess over one grows approximately linearly with $\epsilon$. That is, 
% for sufficiently small $\epsilon$, based on the first order Taylor expansion, we obtain:
% % These observations characterize the behavior of the derived approximation; since it relies on an asymptotic spacing approximation, they do not establish a rigorous finite-sample upper bound on the expected Rashomon-set size.
% \begin{equation}
%     1-\exp\left( -\frac{N\epsilon}{m\sigma} \phi\left(\Phi^{-1}\left(1-\frac{K}{N}\right) \right) \right) = \frac{N\epsilon}{m\sigma} \phi\left(\Phi^{-1}\left(1-\frac{K}{N}\right) \right) + \mathcal{O}(\epsilon^2)
% \end{equation}
% With that, we can rewrite the approximate expected upper bound on the Rashomon-set size as
% \begin{equation}
%     \mathbb E\!\left[|\mathcal R_\epsilon^K|\right] \lesssim\ 1 +\sum_{m=1}^{\min(K,N-K)} \binom Km\binom{N-K}{m} \times \left[ \frac{N\epsilon}{m\sigma} \phi\left(\Phi^{-1}\left(1-\frac{K}{N}\right) \right) + \mathcal{O}(\epsilon^2) \right].
% \end{equation}
% Thus, for sufficiently small tolerances, a first-order expansion of the exponential term shows that the excess over one grows approximately linearly with $\epsilon$.

% \color{red} igual azken hau kenduko nuke \color{black}

\subsection{Preliminary Empirical Insights}
\label{sec:rashomon-empirical-insights}

% To develop intuition and obtain preliminary insights, before delving into
Before analyzing the broader implications of predictive multiplicity in Section~\ref{sec:fairness-analysis}, we begin with a series of illustrative toy experiments. We consider three scenarios: (a) $N=5$, $K=3$; (b) $N=10$, $K=3$; and (c) $N=10$, $K=5$. The corresponding numbers of distinct top-$K$ configurations, given by $\binom{N}{K}$, are 10, 120, and 252, respectively. These values represent the maximum possible size of the Rashomon set in each setting.

We assume that candidate utilities are independently sampled from a zero-mean Gaussian distribution, $y_i\sim\mathcal{N}(0,\sigma^2)$, and vary the standard deviation $\sigma$ to investigate how utility dispersion affects the size of the Rashomon set. When $\sigma$ is small, the utilities are concentrated around their mean, resulting in relatively similar utility values across candidates. As $\sigma$ increases, the utilities become more dispersed, leading to greater differences in candidate utilities. These experiments therefore examine how the size of the Rashomon set, and consequently the multiplicity of top-$K$ configurations, varies with the dispersion of the utility distribution. In addition, we evaluate the derived upper bounds on the Rashomon-set size: the general bound in Equation~\ref{eq:rashomon_size_bound} and the approximate upper bound on its expected size under the i.i.d. Gaussian assumption, given in Equation~\ref{eq:approx-gaussian-rashomon-bound}.

\begin{figure}
     \centering
     \begin{subfigure}[b]{0.47\textwidth}
         \centering
         \includegraphics[width=\textwidth]{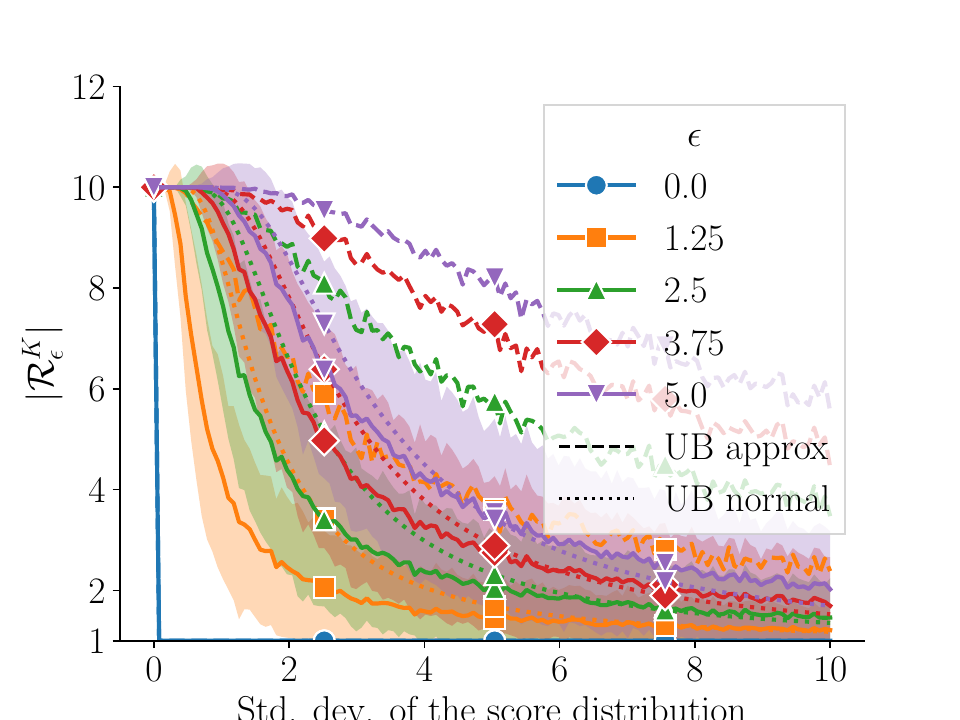}
         \caption{$N=5$, $K=3$.}
         \label{fig:n5-k3}
     \end{subfigure}
     %\hfill
     \begin{subfigure}[b]{0.47\textwidth}
         \centering
         \includegraphics[width=\textwidth]{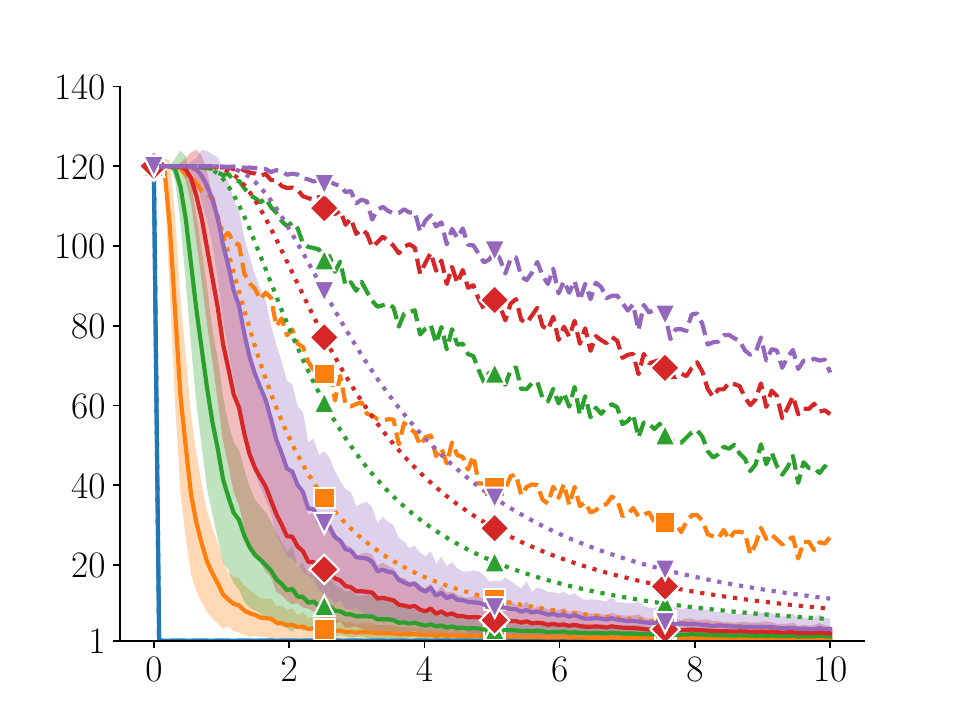}
         \caption{$N=10$, $K=3$.}
         \label{fig:n10-k3}
     \end{subfigure}
     %\hfill
     \begin{subfigure}[b]{0.47\textwidth}
         \centering
         \includegraphics[width=\textwidth]{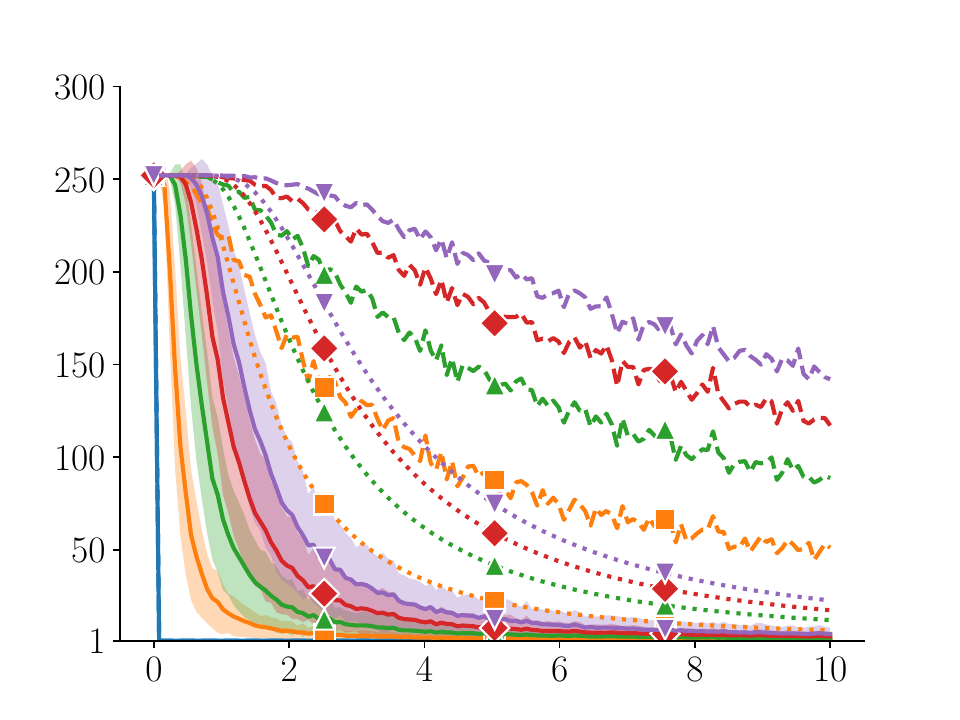}
         \caption{$N=10$, $K=5$.}
         \label{fig:n10_k5}
     \end{subfigure}
        \caption{Rashomon-set size as a function of the standard deviation $\sigma$ for utilities sampled independently from a zero-mean Gaussian distribution, $y_i \sim \mathcal{N}(0, \sigma^2)$. Solid lines represent the average Rashomon-set size across 512 populations per $\sigma$ value, and contours indicate standard deviation. The Rashomon-set size decreases as $\sigma$ increases, reflecting greater separation among candidate utilities. Dashed lines show the general upper bound from Equation~\ref{eq:rashomon_size_bound}, whereas dotted lines denote the upper bound on the expected Rashomon-set size under the i.i.d. Gaussian assumption from Equation~\ref{eq:approx-gaussian-rashomon-bound}. Although the general bound can be loose, accounting for the structure of the utilities' distribution leads to tighter bounds that more accurately capture the observed behavior.}
        \label{fig:toy-exp-bounds}
\end{figure}

The results are presented in Figure~\ref{fig:toy-exp-bounds}. As expected, increasing the dispersion of the candidate utilities (i.e., increasing $\sigma$) leads to a decrease in the size of the Rashomon set. This behavior can be explained by the fact that greater utility differences make candidate exchanges more costly in terms of utility loss. Consequently, fewer alternative top-$K$ configurations can be obtained while remaining within the prescribed utility-loss budget $\epsilon$.
When $\sigma=0$, all candidate utilities are equal, and hence any top-$K$ configuration achieves the same utility. In this case, the Rashomon set attains its maximum possible size, corresponding to the total number of distinct top-$K$ configurations. As predicted, these values are 10, 120, and 252 for the three considered settings, respectively.

Regarding the upper bounds, we observe that the general bound from Equation~\ref{eq:rashomon_size_bound} can be relatively loose, whereas the bound derived under the i.i.d. Gaussian assumption (i.e., Equation~\ref{eq:approx-gaussian-rashomon-bound}) provides a substantially tighter characterization of the observed Rashomon-set size. This highlights the importance of incorporating additional information about the underlying utility distribution when characterizing predictive multiplicity. More broadly, exploiting the structure of a particular problem can yield bounds that are not only tighter, but also more informative about the behavior of the Rashomon set in practice.

\section{The Effect of Multiplicity on the Fairness Guarantees of the Rankings}
\label{sec:fairness-analysis}

The existence of distinct top-$K$ candidate sets that yield nearly identical overall utility opens promising new avenues for studying fairness in ranking systems. When multiple top-$K$ selections achieve comparable utility, their specific compositions can vary drastically, meaning an individual candidate may be selected or excluded depending entirely on which optimal or near-optimal ranking is chosen. This phenomenon illustrates how predictive multiplicity
%—the existence of competing models with equivalent predictive performance but divergent individual outcomes—
can serve as a hidden source of unfairness. Crucially, however, this multiplicity also presents a valuable opportunity: by systematically exploring the candidate sets within this near-optimal space (the Rashomon set), it becomes possible to identify rankings that maximize fairness guarantees without sacrificing overall utility.

Building on these insights, the following sections examine those two complementary facets of predictive multiplicity in the fairness guarantees of rankings. First, we analyze the probability of an individual instance being included in the top-$K$ across all rankings in the Rashomon set, introducing metrics to measure prediction instability and entropy at individual level (Section~\ref{sec:rashomon-flip-prob}). Second, we study the optimal group fairness metrics reachable within the same Rashomon set (Section~\ref{sec:rashomon-group-fairness}).

% With that, in the following sections we study the.... (Section..) and ... (Section ...)

% We study fairness from different perspectives. (a) the arbitrariness (flip probabilities) of the decision, and the possible gap in instability and entropy and (b) the rashomon set as a way to imptove representation and diversity metrics. 

\subsection{The Rashomon Effect as a Source of Unfairness: Quantifying Inclusion Probability}
\label{sec:rashomon-flip-prob}

In this section, we study the new sources of unfairness that may arise as a consequence of the Rashomon Effect. 
For that purpose, we formally define the probability that an individual $j\in[1, N]$ is included within the top-$K$ considering the whole Rashomon set: 
%Based on Equation \ref{eq:rashomon_set}, the inclusion probability is defined as:
\begin{equation}
    p_j^{(\epsilon)} = \frac{|\{ \mathcal{C}_{K} \in \mathcal{R}_\epsilon \; : \; j \in \mathcal{C}_{K} \}|}{|\mathcal{R}_\epsilon|}.
    \label{eq:inclusion-probabilities}
\end{equation}

With this, $p_j^{(\epsilon)}$ quantifies the robustness of membership of the individual $j$ in the top-$K$ across all near-optimal rankings. 
This can be seen as an Rashomon effect-aware notion of instability or uncertainty. If the value of $p_j^{(\epsilon)}$ is close to 1, it means that the instance $j$ is robustly included in the top-$K$, whilst for the values close to 0, it is robustly excluded from the top-$K$. Intermediate values describe unstable or interchangeable membership induced by the Rashomon effect.

The inclusion probability will depend on the tolerance $\epsilon$ chosen to define the Rashomon set and the true scores of the candidates. For instance, starting from the Bayes-optimal ranking, 
where the instances are sorted according to their true utility $y_{\piref(1)} \geq ... \geq y_{\piref(N)}$, 
replacing optimal item $i \leq K$ with non-optimal item $j>K$ in the top-$K$ induces the cost $c_{ij} = y_{\piref(i)} - y_{\piref(j)}$.
% \begin{equation}
%     c_{ij} = y_{\piref(i)} - y_{\piref(j)}
% \end{equation}
Thus, an item $j>K$ can appear in the top-$K$ only if $y_{\piref(K)} - y_{\piref(j)} \leq \epsilon$. Similarly, an item $i<K$ can be excluded only if $y_{\piref(i)} - y_{\piref(K+1)} \leq \epsilon$. Hence, if $y_{\piref(i)} - y_{\piref(K+1)} > \epsilon$, then $p_i^{(\epsilon)}=1$, and if $y_{\piref(K)} - y_{\piref(j)} > \epsilon$, then $p_j^{(\epsilon)}=0$.

Moreover, with the individual belonging probabilities we can define the notion of \textit{selection instability}, $instability(j) = \min (p_j^{(\epsilon)}, 1- p_j^{(\epsilon)}),$
% \begin{equation}
%     instability(j) = \min (p_j^{(\epsilon)}, 1- p_j^{(\epsilon)}),
%     \label{eq:selection-instability}
% \end{equation}
or \textit{entropy}, $H_j = - p_j^{(\epsilon)} \log p_j^{(\epsilon)} - (1- p_j^{(\epsilon)}) \log (1-p_j^{(\epsilon)}).$
% \begin{equation}
%     H_j = - p_j^{(\epsilon)} \log p_j^{(\epsilon)} - (1- p_j^{(\epsilon)}) \log (1-p_j^{(\epsilon)}).
%     \label{eq:selection-entropy}
% \end{equation}

In principle, estimating these inclusion probabilities is challenging, as it requires characterizing the Rashomon set, or equivalently, identifying all top-$K$ configurations that belong to it. Explicitly enumerating these configurations quickly becomes computationally intractable as the number of candidates $N$ and the top-$K$ size increase. Nevertheless, even without fully characterizing the Rashomon set, we can distinguish between candidates whose inclusion probabilities satisfy $p_j^{(\epsilon)}=0$, $p_j^{(\epsilon)}=1$, or $p_j^{(\epsilon)}\in(0,1)$. These cases correspond, respectively, to candidates that are never selected, always selected, or selected only in some of the rankings within the Rashomon set. Examining the demographic characteristics of candidates across these categories can provide valuable insights into the distribution of inclusion instability. In particular, it can help reveal whether the uncertainty surrounding a candidate's inclusion in the top-$K$ is systematically associated with membership in particular demographic groups, thereby shedding light on potential sources of bias in ranking instability. We investigate these potential biases across several real-world ranking problems in Section~\ref{sec:experiments}.

\subsection{The Potential of the Rashomon Effect to Find Less Discriminatory Rankings}
\label{sec:rashomon-group-fairness}

In this section, we investigate how the Rashomon effect, and particularly the Rashomon set, can be leveraged to identify fairer rankings. The formulation of the problem of finding less discriminatory rankings depends on the fairness criterion adopted, which in turn determines the rankings considered fair. As a preliminary analysis, we focus on a representation-based fairness metric. Specifically, we consider a simplified version of representation difference \cite{celis_et_al:LIPIcs:2018:9032}, which is closely related to established legal and policy concepts of equity, including demographic parity \cite{dwork2012fairness} and disparate impact \cite{feldman2015certifying}. Besides, it can be regarded as a particular form of positive action.
This metric measures the difference in representation between the top-$K$ and the whole set of candidates for a given sensitive group $s$. Formally, we describe this metric as:
\begin{equation}
    RD_s(\boldsymbol{\pi}, \mathcal{G}_s) = \frac{|\mathcal{G}_s^{(K)}|}{K} - \frac{|\mathcal{G}_s|}{N}
    \label{eq:representation-difference}
\end{equation}
The ideal fairness state corresponds to $\textrm{RD}_s(\boldsymbol{\pi}, \mathcal{G}_s) = 0$, where the proportion of group $\mathcal{G}_s$ in the top-$K$ outcome perfectly mirrors its proportion in the candidate pool $N$.

We seek to find a candidate set $\hat{\mathcal{C}_K} \in \mathcal{R}_{\epsilon}^K$ that minimizes the representation disparity $\textrm{RD}_s(\boldsymbol{\pi}, \mathcal{G}_s)$ for all the sensitive groups, thereby bringing the group's top-$K$ representation as close to demographic parity as allowed by the budget $\epsilon$. Formally, the ideal target count of group $\mathcal{G}_s$ candidates in top-$K$ for zero representation difference is $K_s^* = K \cdot p_s = K \frac{|\mathcal{G}_s|}{N}$. In discrete candidate spaces, the achievable target integer is $k_s^\text{target} = \lfloor K_s^* \rceil$.
Nevertheless, it is important to note that, when evaluating system fairness across $A$ demographic groups $\{\mathcal{G}_1, \mathcal{G}_2, \dots, \mathcal{G}_A\}$, individual candidate swaps cannot be executed independently. Because candidate selection is a zero-sum constraint ($|\mathcal{C}_K| = K$), including a candidate from group $\mathcal{G}_g$ forces the exclusion of another candidate, inherently altering the representation difference of all other groups.
Let $\mathcal{C}_K \subset \mathcal{C}$ be a selected top-$K$ candidate set ($|\mathcal{C}_K| = K$). The joint representation disparity across $A$ groups is defined as the vector:
\begin{equation}
   {\textrm{\textbf{RD}}}(\boldsymbol{\pi}) = \left[ \textrm{RD}_1(\boldsymbol{\pi}, \mathcal{G}_1), \; \textrm{RD}_2(\boldsymbol{\pi}, \mathcal{G}_2), \; \dots, \; \textrm{RD}_A(\boldsymbol{\pi}, \mathcal{G}_A) \right]^T \in \mathbb{R}^A,
   \label{eq:rd-vector}
\end{equation}
where each component is $\textrm{RD}_g(\boldsymbol{\pi}, \mathcal{G}_g) = \frac{|\mathcal{C}_K \cap \mathcal{G}_g|}{K} - \frac{|\mathcal{G}_g|}{N}$. The overall unfairness of outcome $\mathcal{C}_K$ under aggregate norm $L_p$ ($p \in \{1, 2, \infty\}$) is $f_p(\mathcal{C}_K) = \|{\textrm{\textbf{RD}}}(\mathcal{C}_K)\|_p$.
The $L_1$-norm aligns with a utilitarian fairness framework that optimizes average group equity across the system. In contrast, the $L_\infty$-norm operationalizes an egalitarian or Rawlsian fairness objective by directly minimizing the disparity of the most severely disadvantaged group \cite{martinez2020minimax}. This prevents any single demographic group from bearing a disproportionate burden of representation error. For the remainder of this section, we focus on the $L_1$-norm. Theoretical results on identifying less discriminatory rankings under the $L_\infty$-norm are provided in Appendix~\ref{ap:multi-group-fairness-optim-linf}.

\begin{theorem}[Multi-Group Fair Top-$K$ Optimization in $\mathcal{R}_\epsilon^K$ under $L_1$-Norm]
Let $\alpha_i \in \{0, 1\}$ be a decision variable indicating whether candidate $i \in \mathcal{C}$ is selected into top-$K$ ($\alpha_i = 1$). Let $a_{i, g} = \mathbb{I}(i \in \mathcal{G}_g)$ be the group membership indicator. Finding the globally fairest ranking within the $\epsilon$-level Rashomon set $\mathcal{R}_\epsilon^K$ under $L_1$-norm disparity is equivalent to solving the Binary Integer Linear Program:
\begin{align}
    \min_{\boldsymbol{\alpha}, \boldsymbol{d}^+, \boldsymbol{d}^-} \quad & \sum_{g=1}^A \left( d_g^+ + d_g^- \right) \label{obj:multi_fairness} \\
    \text{s.t.} \quad & \sum_{i=1}^N \alpha_i = K, \label{cons:top_k} \\
    & \sum_{i \in \setref} y_i (1 - \alpha_i) - \sum_{j \notin \setref} y_j \alpha_j \le \epsilon, \label{cons:rashomon} \\
    & \frac{1}{K} \sum_{i=1}^N a_{i, g} \alpha_i - \frac{|\mathcal{G}_g|}{N} = d_g^+ - d_g^-, \quad \forall g \in \{1, \dots, A\}, \label{cons:disparity_slack} \\
    & \alpha_i \in \{0, 1\}, \quad \forall i \in \{1, \dots, N\}, \label{cons:binary} \\
    & d_g^+, d_g^- \ge 0, \quad \forall g \in \{1, \dots, A\}. \label{cons:slack_nonneg}
\end{align}
In this formulation, constraint \eqref{cons:top_k} enforces cardinality, \eqref{cons:rashomon} restricts outcomes to the $\epsilon$-level Rashomon set, and \eqref{cons:disparity_slack} decomposes absolute group representation error using non-negative slack variables $d_g^+$ and $d_g^-$.
\label{th:multi-group-fairness-optim}
\end{theorem}

The proof of Theorem \ref{th:multi-group-fairness-optim} can be found in Appendix \ref{ap:proof-fairness-optim}. The optimization problem from Theorem \ref{th:multi-group-fairness-optim} is structurally equivalent to the Constrained Multidimensional Knapsack Problem (MKP) \cite{kellerer2004multidimensional}, and therefore, 
the problem is NP-hard. Consequently, finding an exact global optimum for large candidate pools ($N$) or intersecting group structures requires Mixed-Integer Linear Programming (MILP) \cite{vielma2015mixed} exact solvers using Branch-and-Cut algorithms (e.g., Gurobi). When the total candidate pool $N$ is moderate (e.g., $N \le 10^4$) the latter solvers routinely identify the globally optimal ranking within milliseconds to seconds. However, for large-scale applications where exact solving becomes computationally intractable, polynomial-time approximations can be obtained by solving the continuous LP relaxation followed by dependent randomized rounding, or by applying Lagrangian relaxation to decouple the group fairness constraints. Furthermore, heuristic methods can provide computationally efficient alternatives, at the expense of a general guarantee of global optimality.

\section{Empirical Analysis on Real-World Ranking Problems}
\label{sec:experiments}

In this section, we conduct extensive experiments on several real-world ranking problems to investigate the fairness implications of predictive multiplicity in ranking problems. As discussed in the previous section, we consider two complementary perspectives: (i) ranking multiplicity as a potential source of unfairness, and (ii) the characterization of the Rashomon set as an opportunity to identify less discriminatory rankings without sacrificing utility. More specifically, we examine whether ranking instability, particularly uncertainty regarding candidates' inclusion in or exclusion from the top-$K$, exhibits systematic disparities across demographic groups. Furthermore, we also evaluate the effectiveness of the results in Theorem~\ref{th:multi-group-fairness-optim} in identifying the least discriminatory rankings within the Rashomon set.

\subsection{Experimental Setting}

\subsubsection{Datasets} We consider a diverse set of real-world datasets spanning a variety of tasks and demographic characteristics: (a) \texttt{LawSAT}~\citep{kusner2017counterfactual}, law school entrance scores; (b) \texttt{HMDA}~\citep{sourceHDMA}, mortgage data for the state of New York in 2025, using loan amount as the score attribute; (c) \texttt{SAT}~\citep{sourceSAT}, college performance data in the US, using verbal and mathematics SAT scores; (d) \texttt{COMPAS}~\citep{sourceCompas}, recidivism risk scores produced by the COMPAS algorithm used across several US states; and (e) \texttt{Astronauts}~\citep{sourceAstronauts}, flight hours of NASA astronauts since 1959. We treat gender and ethnicity as sensitive attributes in all datasets, except for \texttt{Astronauts} and \texttt{SAT}, which include only gender. A more detailed characterization of these datasets and their properties is provided in Appendices~\ref{ap:datasets} and~\ref{ap:instability}. For all datasets, we normalized the score attributes to the $[0, 1]$ interval. Additionally, we oriented the score values such that higher values consistently indicate better outcomes and binarized the sensitive attributes.\footnote{We acknowledge that sensitive attributes such as gender and ethnicity are far more complex than discrete categories. This binarization serves only to facilitate visualization and a preliminary understanding of fairness-related issues in real-world ranking problems. Please refer to the ``Ethics and Privacy Statement'' at the end of this document for further discussion.}

\begin{wrapfigure}{r}{0.35\textwidth}
  \vspace{-\baselineskip} 
  \begin{center}
    \includegraphics[width=0.33\textwidth]{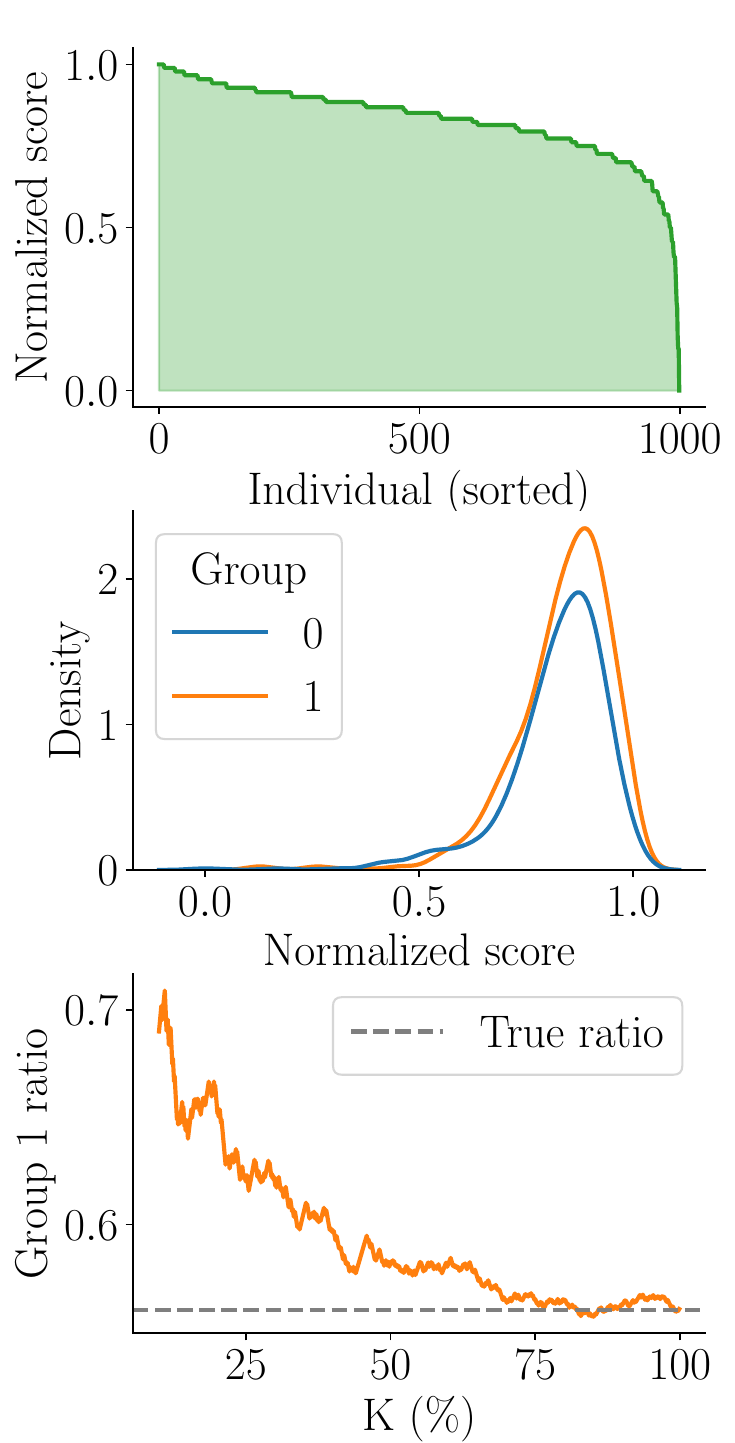}
  \end{center}
  \caption{(Top to bottom) Distribution of true scores across the full dataset (1) and the sensitive groups (2), and the proportion of the majority group in the top-$K$ of the Bayes-optimal ranking for different values of $K$, compared with its proportion in the overall dataset (3). \texttt{Law SAT} dataset, with \textit{gender} as the sensitive attribute.}
  \vskip -0.4in
  \label{fig:dataset-ch-scores}
\end{wrapfigure}

\subsubsection{Optimization} To find less discriminatory algorithms, we solve the optimization problem from Theorem~\ref{th:multi-group-fairness-optim} using the \texttt{PuLP} \citep{mitchell2011pulp} the Python package and the \texttt{MOSEK} solver \cite{mosek}. Due to computational resource constraints in the optimization experiments (Figure~\ref{fig:lp-results}), for datasets exceeding 1,000 entries, we use a random sample of 1,000 individuals. Hardware details are given in Appendix~\ref{ap:hardware}

\subsection{Results}

We begin our empirical study by characterizing the datasets to identify potential sources of bias in the ranking problems under consideration. Specifically, we examine the overall distribution of true scores, their distributions across sensitive groups, and the proportion of the majority group in the top-$K$ of the Bayes-optimal ranking for different values of $K$. Figure~\ref{fig:dataset-ch-scores} presents these analyses for the \texttt{LawSAT} dataset. The figure shows that, despite having similar score distributions, male candidates are overrepresented in the top-$K$ relative to their proportion in the overall dataset, particularly for smaller values of $K$. This disparity diminishes as $K$ increases, since a larger share of candidates is included in the ranking. These results suggest that fairness disparities may become more pronounced in highly competitive ranking settings, where only a small fraction of candidates are selected.
Similar trends leading to the same conclusions across the remaining datasets are provided in Appendix~\ref{ap:score-distributions}.

\begin{figure}
    \centering
    \includegraphics[width=\linewidth]{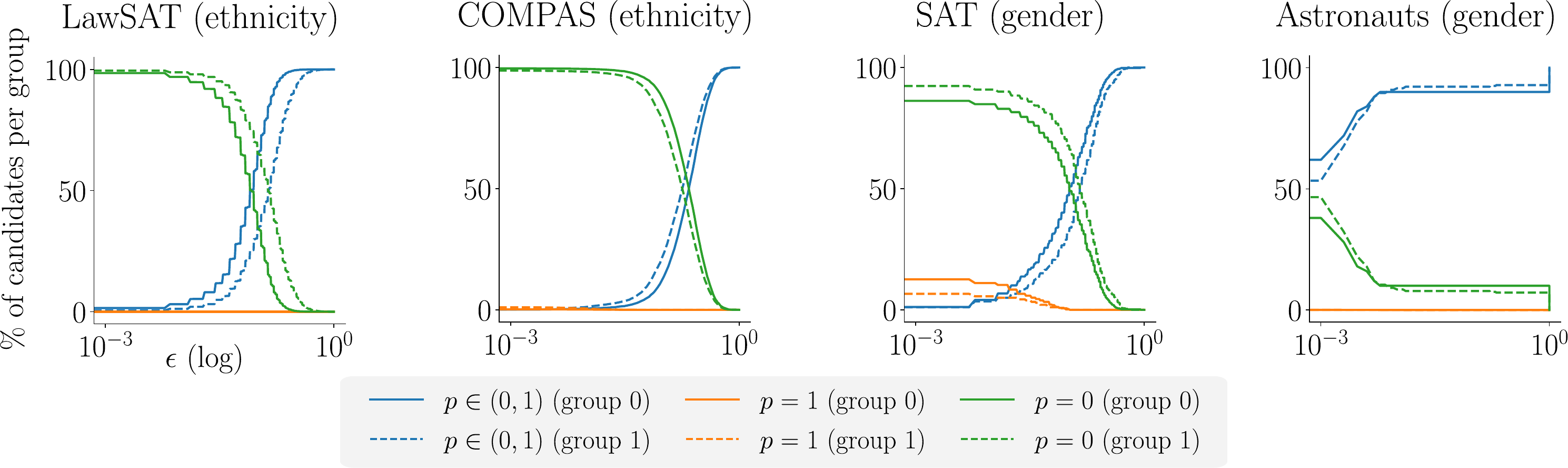}
        \caption{Probability of inclusion ($p_j^{(\epsilon)}$) in the top-$K$ across different values of $\epsilon$. We distinguish three cases: $p_j^{(\epsilon)}=0$, indicating candidates who are never selected; $p_j^{(\epsilon)}=1$, indicating candidates who are always selected; and $p_j^{(\epsilon)}\in(0,1)$, indicating candidates who are selected in some, but not all, rankings within the Rashomon set. The figure further disseminates the distribution of these cases differs across sensitive groups. In these experiments we consider $K=100$. Results for the rest of the datasets are deferred to Appendix~\ref{ap:instability}.}
        %\vskip -2mm
    \label{fig:placeholder}
\end{figure}

\begin{figure}
    \centering
    \includegraphics[width=\linewidth]{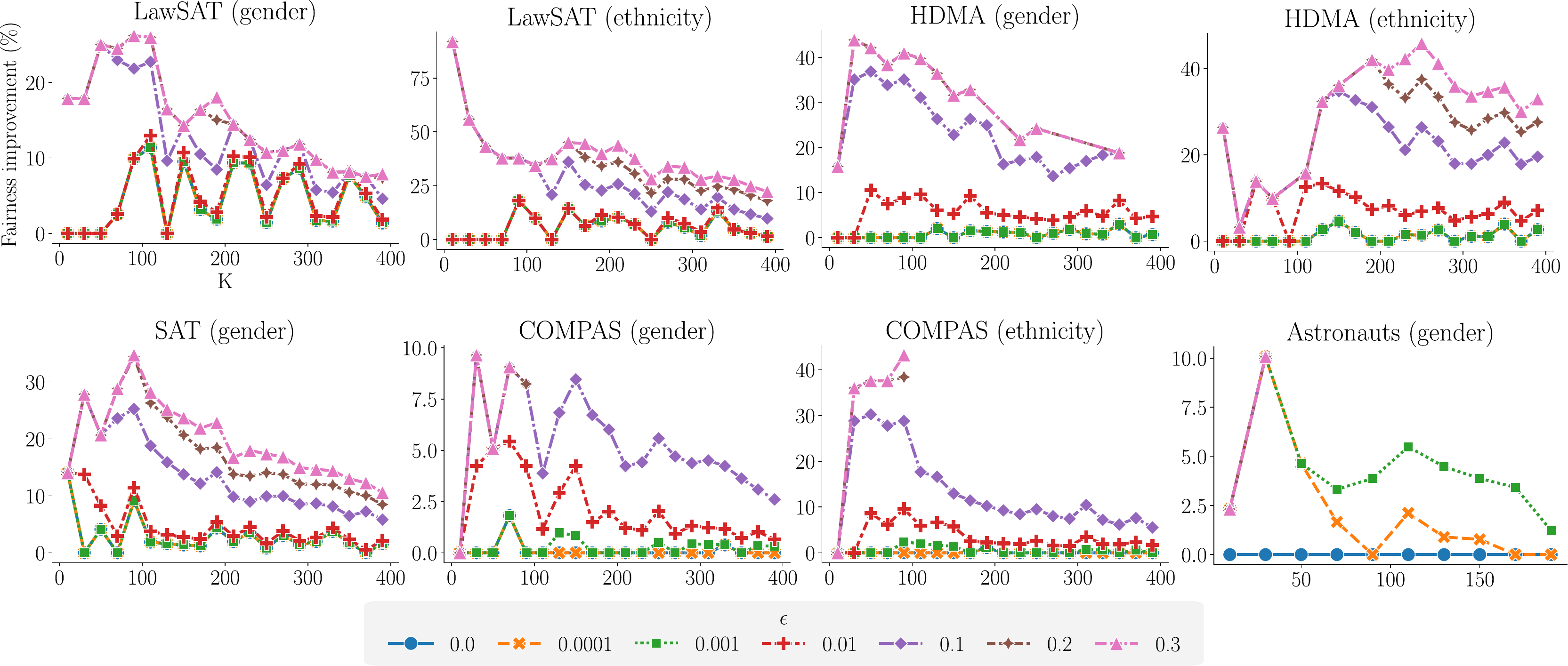}
    \caption{Fairness improvements achieved by solving the optimization problem in Theorem~\ref{th:multi-group-fairness-optim} to identify less discriminatory rankings across different top-$K$ sizes and real-world datasets, for different values of $\epsilon$. Allowing a larger utility-loss tolerance $\epsilon$ generally yields greater fairness improvements. Notably, substantial improvements can also be achieved in many cases with $\epsilon=0$, demonstrating that, by accounting for the Rashomon effect, less discriminatory rankings may be obtained without sacrificing utility. Missing data points for some $\epsilon = 0.3$ lines reflect instances where the solver reached the 2-minute time limit.}
    \label{fig:lp-results}
    %\vskip -0.2in
\end{figure}

In Figure~\ref{fig:placeholder} we examine how the distribution of inclusion probabilities discussed in Section~\ref{sec:rashomon-flip-prob} varies across sensitive groups and ranking tasks. In most datasets, for small values of $\epsilon$, the majority of candidates have zero probability of being included in the top-$K$ ($p_j^{(\epsilon)}=0$), while only a few are guaranteed inclusion ($p_j^{(\epsilon)}=1$) or exhibit inclusion uncertainty ($p_j^{(\epsilon)}\in(0,1)$). As $\epsilon$ increases, the proportion of candidates with zero inclusion probability decreases, while that of candidates with intermediate inclusion probabilities increases, reflecting greater variability in top-$K$ membership across the Rashomon set.
However, the \texttt{Astronauts} dataset exhibits a different pattern: for small values of $\epsilon$, the proportions of candidates with $p_j^{(\epsilon)}=0$ and $p_j^{(\epsilon)}\in(0,1)$ are comparable, while very few candidates are guaranteed inclusion. This behavior can be attributed to the smaller differences in true scores between candidates, particularly among those with high scores.  As $\epsilon$ increases, the proportion of candidates with zero inclusion probability decreases markedly, accompanied by a corresponding increase in the proportion with intermediate inclusion probabilities.

We further observe consistent group-level disparities in the \texttt{LawSAT} and \texttt{SAT} datasets: Group 0 contains more candidates who are never selected ($p_j^{(\epsilon)}=0$), whereas Group 1 contains more candidates who are either always selected ($p_j^{(\epsilon)}=1$) or have uncertain inclusion ($p_j^{(\epsilon)}\in(0,1)$). A similar pattern emerges in the \texttt{COMPAS} dataset, but with the groups reversed. In contrast, the direction of these disparities in the \texttt{Astronauts} dataset varies with $\epsilon$. Overall, these findings suggest that ranking instability, as well as guaranteed inclusion and exclusion from the top-$K$, are unevenly distributed across sensitive groups, highlighting ranking multiplicity as a potential source of group-level disparities.

Furthermore, Figure~\ref{fig:lp-results} illustrates the improvements obtained in ranking fairness by considering the Rashomon effect through the results from Theorem~\ref{th:multi-group-fairness-optim}. The fairness guarantees of the ranking are measured by the $L_1$-norm of the vector of the group-wise representation-difference metric from Equation~\ref{eq:representation-difference}. These results show that accounting for the Rashomon effect provides a promising approach to identifying less discriminatory rankings while incurring little or no utility loss. As expected, allowing a larger utility-loss tolerance $\epsilon$ generally leads to greater fairness improvements, since a larger Rashomon set offers a broader range of alternative top-$K$ configurations from which to select a fairer ranking. Importantly, fairness improvements are also observed when $\epsilon=0$, indicating that less discriminatory rankings can sometimes be identified without any loss in utility. 

Besides, we observe that fairness improvements are generally more pronounced for smaller values of $K$. As discussed in the previous empirical analysis, fairness disparities tend to be more severe in highly competitive ranking settings, where only a small fraction of candidates is selected. Our results therefore suggest that leveraging the Rashomon effect can yield substantial fairness improvements without sacrificing utility, even in such competitive settings. Aligning with the intuition built from Figure~\ref{fig:dataset-ch-scores} and Appendix~\ref{ap:instability}, improvements tend to be smaller for larger values of $K$, as including a greater proportion of the candidate population makes the demographic composition of the top-$K$ more likely to reflect that of the overall population, thereby reducing initial representation disparities and leaving less room for further improvement.

\section{Conclusions}
\label{sec:conclusions}

In this work, we formalize the Rashomon effect in score-based ranking problems by introducing the top-$K$ Rashomon set, which comprises alternative top-$K$ configurations whose utility is equal or sufficiently close to that of the Bayes-optimal configuration. We characterize predictive multiplicity through the size of this set, and derive theoretical results that illuminate its extent under different assumptions about the utility distribution. We then examine the implications of predictive multiplicity for fairness. On the one hand, it may be regarded as a new source of unfairness, as candidates’ inclusion in the top-$K$ can vary across equally or nearly equally utility-optimal configurations. On the other hand, the alternative configurations it provides can be leveraged to identify less discriminatory rankings with little or no loss in utility.

Experiments on illustrative toy problems and real-world datasets demonstrate the practical relevance of our framework and support our theoretical findings. Overall, our results highlight the dual role of predictive multiplicity in ranking: it can expose potential sources of unfairness arising from ranking instability, while also offering a principled means of improving fairness without necessarily compromising utility.

We acknowledge that Section~\ref{sec:rashomon-group-fairness} focuses on improving fairness guarantees with respect to a single fairness metric. While this allows us to highlight the potential of leveraging the Rashomon effect to identify less discriminatory rankings without sacrificing utility, extending the corresponding optimization formulations to a broader range of fairness criteria is beyond the scope of this work. Future research could extend our framework to incorporate additional fairness metrics and investigate how the choice of criterion affects the selection of fair rankings within the Rashomon set.

Several other interesting directions for future work remain open. First, our analysis considers rankings defined directly through candidate utility scores, without explicitly modeling the learning process used to obtain these scores. In practice, rankings are often induced by selecting a scoring function from a hypothesis class or model family \cite{burges2005learning, xia2008listwise}. Such restrictions may limit the set of attainable rankings and, consequently, alter the structure and size of the corresponding Rashomon set. Incorporating these modeling constraints would provide a more comprehensive understanding of predictive multiplicity in learning-to-rank systems \cite{burges2005learning, xia2008listwise}. Finally, we consider deterministic rankings. Extending our framework to stochastic rankings \cite{lattimore2018toprank, bruch2020stochastic, diaz2020evaluating} could reveal additional forms of multiplicity and instability, while opening new opportunities for fairness-aware ranking under uncertainty.

\section*{Ethics and Privacy Statement}

This work identifies ranking multiplicity as a potential source of unfairness and investigates how it can be leveraged to identify less discriminatory rankings without sacrificing utility. By examining disparities in ranking instability and exploring fairness-aware selection among similarly performing rankings, our approach may contribute to more equitable decision-making in high-stakes applications. However, we acknowledge that concepts such as ``bias'' and ``fairness'' are normative, context-dependent, and shaped by broader social and institutional structures. Similarly, demographic groups are socially constructed categories with fluid and contested boundaries \cite{hanna2020towards}, and demographic labeling practices, particularly those concerning race, may reinforce problematic assumptions and social hierarchies \cite{doh2025position}. Nevertheless, analyzing disparities across demographic attributes remains important for identifying and mitigating unequal model behavior, including disparities affecting intersectional identities.

Our approach should therefore be understood as a tool for auditing and mitigating specific forms of algorithmic disparity, rather than as a comprehensive or universally applicable definition of fairness. The choice of fairness criteria, sensitive attributes, and acceptable utility trade-offs should be informed by the context of deployment and the perspectives of affected stakeholders. Moreover, selecting a less discriminatory ranking from the Rashomon set does not necessarily eliminate underlying societal biases in the data or the decision-making process. The use of sensitive attributes for fairness assessment also raises privacy concerns, particularly when such attributes are incomplete, or inferred. Their use should therefore be subject to appropriate data protection safeguards, purpose limitation, and applicable legal and ethical requirements. Finally, although our findings may support more equitable ranking decisions, the proposed methods could also be misused to justify existing ranking practices or provide a false sense of fairness if their assumptions and limitations are overlooked. We therefore emphasize the importance of contextual evaluation, transparency, and human oversight when applying these methods in consequential settings.

\section*{Acknowledgments}

We are grateful to Xabier Benavides for reading preliminary versions of this paper and providing insightful feedback.

This work is supported by the Basque Government under grant IT1504-22 and through the BERC 2022-2025 program; and by the Spanish Ministry of Science and Innovation under the grant PID2022-137442NB-I00.
% and through BCAM Severo Ochoa accreditation CEX2021-
% 001142-S / MICIN / AEI / 10.13039/501100011033.
Mikel Malagón acknowledges a predoctoral grant from the Spanish MICIU/AEI with code PREP2022-000309, associated with the research project PID2022-137442NB-I00 funded by the Spanish MICIU / AEI / 10.13039 / 501100011033 and FEDER, EU.

\bibliography{ref}
\bibliographystyle{plainnat}

\appendix

\section{Related Works (A Longer Version)}
\label{ap:related-works-long}

This work lies at the intersection of research on the Rashomon effect, predictive multiplicity, and the new potential sources of unfairness arising from these phenomena, as well as the research domain of assessing fairness in ranking systems. 
Existing literature on predictive multiplicity and algorithmic fairness (either in isolation or jointly) has predominantly focused on classification and prediction settings. Although fairness in ranking has received increasing attention, it remains less explored than fairness in classification and prediction. More importantly, the effect of predictive multiplicity in ranking remains largely unexplored. Moreover, and by extention, little is known about how the existence of multiple similarly performing rankings may affect the fairness properties and guarantees of the resulting ranked outcomes. Our work addresses this gap by studying predictive multiplicity in ranking and investigating its implications for algorithmic fairness.

In the following sections, we provide an overview of the literature on predictive multiplicity in classification and fairness-enhancing methods for ranking systems. We then position our work within these research areas, highlighting the existing gaps and, in particular, the limited attention given to the intersection of predictive multiplicity, ranking, and algorithmic fairness.

\subsection{Preditive Multiplicity in Classification and Its Implications on Fairness}

Machine learning algorithms are often designed to identify the best-fitting model for a given prediction task. However, several studies have shown that prediction problems can exhibit \textit{predictive multiplicity}: the existence of multiple competing models that achieve nearly equivalent predictive performance while potentially offering different explanations of the data-generating process \cite{marx2020predictive, watson2023predictive}. This phenomenon, which Breiman described as the \textit{Rashomon effect}, raises an important question: when multiple models perform similarly well but differ in their predictions or underlying explanations, how should we determine which model is most appropriate?

Predictive multiplicity has important implications for both model selection and algorithmic fairness. In many deployment scenarios, models are selected primarily on the basis of predictive accuracy. However, when several models achieve comparable accuracy, this criterion alone is insufficient to identify the model best suited to the task. In this sense, accuracy-based model selection is an underspecified process \cite{black2022model}: unless additional considerations are explicitly incorporated into model development, there is little reason to expect that the selected model will be the one that best satisfies other objectives, such as minimizing disparate impact or improving interpretability and robustness.

These concerns are particularly important in high-stakes applications, such as lending and recidivism prediction, where algorithmic decisions can significantly affect people's lives \cite{binns2018s}. When equally accurate models produce conflicting predictions, the choice of model can have substantial consequences for individuals and demographic groups. Predictive multiplicity therefore introduces challenges across several stages of the machine learning pipeline, including model selection, model validation, and post-hoc explanation, while raising questions about the fairness guarantees of automated decision-making systems.

At the same time, predictive multiplicity presents an opportunity \cite{black2022model}. The existence of multiple models with comparable predictive performance gives model developers greater flexibility to account for objectives beyond accuracy. Rather than sacrificing predictive performance to achieve other desirable properties, developers can potentially select among equally accurate models to prioritize values such as fairness, robustness, and interpretability. From this perspective, multiplicity provides an opportunity to improve the alignment between predictive models and the broader objectives of the applications in which they are deployed.

To date, research on predictive multiplicity and its relationship to fairness constraints has focused primarily on classification problems. However, many real-world automated decision-making systems involve ranking or prioritization rather than assigning individuals to discrete classes. For example, social welfare agencies may need to prioritize applicants for scarce resources, such as social housing or financial assistance. Such applications commonly follow a score-based ranking paradigm \cite{zehlike2022fairness}, in which a scoring function assigns scores that determine the relative priority of candidates, with resources allocated to those ranked highest.

In these settings, predictive multiplicity may have implications beyond differences in individual predictions: equally accurate scoring functions can induce different rankings and, consequently, lead to different resource-allocation outcomes across demographic groups. Understanding how multiplicity interacts with fairness in ranking is therefore essential for determining whether, and how, the flexibility afforded by multiple equally accurate models can be used to promote fairer allocation decisions.

\subsection{Fairness in Score-based Ranking Systems} 

The literature on algorithmic fairness has predominantly focused on classification and prediction tasks \cite{pessach2022review}. However, many real-world applications of automated decision-making are inherently ranking or prioritization problems. For example, social welfare agencies may need to prioritize applicants for limited resources, such as social housing or financial assistance. More generally, these applications follow a score-based ranking paradigm \cite{zehlike2022fairness}, in which a scoring function determines the relative priority of candidates. For instance, in such settings, an algorithm can assign each applicant a priority score based on characteristics such as income, household composition, or housing conditions, and rank applicants accordingly. Resources can then be allocated to the highest-ranked individuals. This makes fairness in ranking particularly important, as differences between similarly performing scoring functions may lead to different rankings and, consequently, different allocation outcomes across demographic groups.

\subsubsection{Assesing Fairness Guarantees}

In ranking-based systems, several distinct types of bias can emerge. Pre-existing bias exists independently of the algorithm, originating from broader societal factors such as systemic racial or socioeconomic inequalities. Technical bias arises from algorithmic constraints and system design, such as position bias in search results. Finally, emergent bias occurs during real-world use—often when a system is deployed to a user demographic different from the one originally intended, or when societal norms shift over time. Left unaddressed, these biases can reinforce one another, creating damaging feedback loops.

To address these challenges, the scientific community has proposed various metrics to evaluate fairness guarantees in ranked outcomes, using them as guiding objectives when developing bias mitigation methods. Generally, these metrics focus on the final ranking and assume that the input scores used to generate the ranking already encode pre-existing bias.

The first category of metrics enforces representation-based fairness, ensuring that protected groups are equitably represented in the final ranked outcomes. For example, metrics evaluating proportional representation \cite{yang2017measuring} measure the disparity between group proportions within the top-$k$ positions compared to the overall candidate pool. Similarly, diversity-based metrics \cite{celis_et_al:LIPIcs:2018:9032, stoyanovich2018online} establish explicit upper and lower bounds on group representation.

In contrast, a second category moves beyond simple representation to consider the utility loss experienced by each group, aiming to balance this loss equitably across demographics. For instance, the IGF-Ratio \cite{yang_ijcai2019p836} quantifies within-group utility by comparing the score of the highest-ranking skipped candidate to that of the lowest-ranking selected candidate.

Additionally, to account for position bias—where higher-ranked individuals receive significantly more attention—exposure-based measures quantify whether exposure is fairly distributed across groups. These measures evaluate either whether exposure is equal across groups \cite{Singh2018fairness, singh2019policy} or whether group exposure reflects group proportions in the population \cite{Sapiezynski2019quantifying, raj2022measuring}.

Pairwise measures consider pairs of candidates from different groups and examine whether, on average, any group is systematically disadvantaged—that is, whether candidates from a particular group are more likely to be ranked behind candidates from another group than vice versa \cite{beutel2019fairness, narasimhan2020pairwise}.

Expanding on individual considerations, counterfactual fairness \cite{yang_et_al:LIPIcs.FORC.2021.7} assesses whether an outcome for a specific candidate would change if their sensitive attribute were altered (i.e., comparing them to a counterfactual "twin" from a privileged group). A ranking is deemed counterfactually fair only if every candidate's outcome remains unchanged under this counterfactual intervention.

For a comprehensive review of these metrics, we refer the reader to \citet{schumacher2025properties}. To operationalize these metrics, researchers have developed various fairness-enhancing interventions.

\subsubsection{Fairness-enhancing Interventions}

In the literature, methods designed to enhance the fairness guarantees of rankings are typically categorized by the stage at which they intervene: modifying the input score distribution, adjusting the scoring function itself, or re-ranking the final output.

Methods that intervene on the score distribution \cite{kleinberg_et_al:LIPIcs:2018:8323, celis2020interventions, yang_et_al:LIPIcs.FORC.2021.7} aim to mitigate disparities in candidate scores. For instance, \citet{kleinberg_et_al:LIPIcs:2018:8323} and \citet{celis2020interventions} assume that the perceived scores of underprivileged candidates are systematically downgraded relative to their true scores, and they seek rankings that maximize true utility rather than perceived utility. Additionally, \citet{yang_et_al:LIPIcs.FORC.2021.7} use a structural causal model (SCM) to capture the causal effects of sensitive attributes on other variables, generating fair rankings by removing these effects via model-based counterfactuals.

On the other hand, methods that intervene on the ranking function \cite{asudeh2019designing} identify a function similar to the original input function that produces a ranked outcome satisfying specified fairness criteria. For example, \citet{asudeh2019designing} focus on linear functions, automating the search for a ranking function that remains close to the original function—measured by the angular distance between their weight vectors—while remaining fair according to a designated fairness oracle. They employ a combinatorial geometry approach to explore the search space efficiently.

Finally, methods that intervene on the ranked outcome impose constraints to guarantee specific levels of diversity or representation among the top-$K$ items as a set or within every prefix of the ranking list \cite{yang2017measuring, celis_et_al:LIPIcs:2018:9032, yang_ijcai2019p836}. For instance, \citet{yang2017measuring} propose an approach adapted from the fair representation learning framework of \citet{zemel2013learning} to ensure balanced group representation in ranked outcomes. Additionally, \citet{celis_et_al:LIPIcs:2018:9032} introduce a method designed to maximize overall utility with respect to an original metric while satisfying strict lower and upper bounds on group representation. Furthermore, \citet{yang_ijcai2019p836} focus on balancing utility loss across groups, ensuring that top-scoring individuals from historically disadvantaged groups have a fair opportunity to appear in the top-$K$, based on the premise that scores are more informative within groups than across them.

Existing approaches to enhancing fairness in score-based ranking problems ultimately focus on identifying a single scoring function or final ranking that satisfies a desired fairness criterion. Consequently, they generally do not account for the possibility that multiple similarly performing scoring functions—and therefore multiple different rankings—may exist. This possibility, captured by the Rashomon effect or predictive multiplicity, is particularly relevant in ranking settings, where alternative models can lead to different top-ranked candidates and, consequently, different allocation outcomes across groups. In this work, we explicitly study this phenomenon in ranking problems, characterizing the multiplicity of high-performing rankings and examining its implications for algorithmic fairness.

\section{Proofs from Section \ref{sec:rashomon_ranking}}

\subsection{Proof of Theorem \ref{th:rashomon_set_size}}
\label{ap:proof_rashomon_size}

\begin{proof}[Proof of Theorem \ref{th:rashomon_set_size}]
Any feasible top-$K$ set $\mathcal{C}_{K} \in \mathcal{R}_\epsilon$ differs from the optimal set
$\setref$ by replacing a subset of items. Let $m = |\setref \setminus \mathcal{C}_{K}|$ denote the number of optimal items removed; since $|\mathcal{C}_{K}| = K$, exactly $m$ non-optimal items must be added, i.e., $|\mathcal{C}_{K} \setminus \setref| = m$. Note that $m \leq \min\{K, N-K\}$, since the maximum number of distinct elements between two top-$K$ sets is at most $\min\{K, N-K\}$.

Each replacement necessarily removes an item from $\setref$ and replaces it with an item from $[N]\setminus \setref$. By the ordering $y_{\piref(1)} \geq ... \geq y_{\piref(N)}$, the smallest possible utility loss from a single such boundary crossing is $\gamma_K = y_{\piref(K)} - y_{\piref(K+1)}$. Hence, each replacement pair contributes at least $\gamma_K$ to the utility loss, implying
\begin{equation*}
    m \cdot \gamma_K \le \epsilon,
    \qquad \text{so that} \qquad
    m \le M = \min \{\lfloor \epsilon / \gamma_K \rfloor, K, N-K \} .
\end{equation*}

For a fixed value of $m$, the number of ways to choose which optimal items are removed is $\binom{K}{m}$, and the number of ways to choose which non-optimal items are added is $\binom{N-K}{m}$. Therefore, the number of top-$K$ sets at distance $m$ from $\setref$ is at most $\binom{K}{m}\binom{N-K}{m}$.
Summing over all feasible values of $m$ yields
\begin{equation}
    |\mathcal{R}_\epsilon^K|
\le
\sum_{m=0}^{M}
\binom{K}{m}\binom{N-K}{m}.
\end{equation}

\end{proof}

\subsection{Proof of Proposition~\ref{prop:rashomon_size_param}}
\label{ap:proof_param_size}

\begin{proof}[Proof of Proposition~\ref{prop:rashomon_size_param}]
Let us analyze each regime separately.

\paragraph{(1) Fixed $K$ and $M$.}
For each fixed $m \le M$, we use the standard bounds
\begin{equation*}
    \binom{K}{m} \le K^m, \qquad \binom{N-K}{m} \le (N-K)^m \le N^m.
\end{equation*}

Hence,
\begin{equation*}
    \binom{K}{m}\binom{N-K}{m} \le K^m N^m = O(N^m),
\end{equation*}
and summing over $m \le M$ yields $|\mathcal{R}_\epsilon^K| = O(N^M).$

\paragraph{(2) Dense regime $K = \alpha N$.}
Let $K = \alpha N$ with $\alpha \in (0,1)$. Using Stirling-type approximations for fixed $m$,
\begin{equation*}
    \binom{\alpha N}{m} \sim \frac{(\alpha N)^m}{m!},
\qquad
\binom{(1-\alpha)N}{m} \sim \frac{((1-\alpha)N)^m}{m!}.
\end{equation*}

Thus,
\begin{equation*}
    \binom{K}{m}\binom{N-K}{m}
\sim
\frac{\left(\alpha(1-\alpha)N^2\right)^m}{(m!)^2}.
\end{equation*}

Summing over $m \le M$ yields $|\mathcal{R}_\epsilon^K| = O\!\left(N^{2M}\right),$
with multiplicative constants depending on $\alpha$ and $m$.

\paragraph{(3) Degenerate case, $\gamma_K = 0$ for $\forall K \leq N$.}
Since $M = \min (\lfloor \epsilon / \gamma_K \rfloor, K)$ and $\lfloor \epsilon / \gamma_K \rfloor \rightarrow \infty$, we have $M=K$. Thus the inequality becomes:
\begin{equation*}
    |\mathcal{R}_\epsilon^K|
\le
\sum_{m=0}^{K}
\binom{K}{m}\binom{N-K}{m}
\end{equation*}

As $\binom{K}{m} = \binom{K}{K-m}$ we get:
\begin{equation*}
    |\mathcal{R}_\epsilon^K|
\le
\sum_{m=0}^{K}
\binom{K}{K-m}\binom{N-K}{m},
\end{equation*}

and due to Vandermonde's identity, we obtain
\begin{equation*}
    |\mathcal{R}_\epsilon^K|
\le \binom{N}{K}. 
\end{equation*}

Furthermore, since $\gamma_K = 0$ for $\forall K \leq N$, every subset $\mathcal{C}_K$ becomes admissible, implying 
\begin{equation*}
    |\mathcal{R}_\epsilon^K| = \binom{N}{K}. 
\end{equation*}

\paragraph{(4) Vanishing tolerance $\epsilon \to 0$.}
If $\epsilon \to 0$, then $M=0$, so only the term $m=0$ contributes:
\begin{equation*}
    \binom{K}{0}\binom{N-K}{0} = 1.
\end{equation*}
Thus $|\mathcal{R}_\epsilon^K| = 1$.
\end{proof}

\subsection{Proof of Theorem \ref{th:boundary-gap-continuous}}
\label{ap:boundary-gap-continuous}

\begin{proof}[Proof of Theorem \ref{th:boundary-gap-continuous}]
    
Let $U=Y_{[K+1]}$ and $V=Y_{[K]}$, so that $\gamma_K=V-U$. The joint density of these adjacent descending order statistics is

$$
\begin{aligned}
f_{U,V}(y,t)
&=\frac{N!}{(N-K-1)!(K-1)!}
F_Y(y)^{N-K-1}\\
&\quad\cdot
\bigl(1-F_Y(t)\bigr)^{K-1}
f_Y(y)f_Y(t),
\qquad y<t.
\end{aligned}
$$

Therefore,

$$
\begin{aligned}
\mathbb{E}[\gamma_K]
&=\int_{-\infty}^{\infty}\int_y^\infty
(t-y)f_{U,V}(y,t)\,dt\,dy\\
&=\frac{N!}{(N-K-1)!(K-1)!}
\int_{-\infty}^{\infty}
F_Y(y)^{N-K-1}f_Y(y)\\
&\quad\cdot
\left(
\int_y^\infty
(t-y)\bigl(1-F_Y(t)\bigr)^{K-1}
f_Y(t)\,dt
\right)dy.
\end{aligned}
$$

Applying integration by parts to the inner integral yields

$$
\begin{aligned}
&\int_y^\infty
(t-y)\bigl(1-F_Y(t)\bigr)^{K-1}f_Y(t)\,dt\\
&\qquad=\frac{1}{K}
\int_y^\infty
\bigl(1-F_Y(t)\bigr)^K\,dt,
\end{aligned}
$$

where the boundary term vanishes under the usual integrability conditions. Substituting this identity and using

$$
\frac{N!}{K!(N-K-1)!}
=N\binom{N-1}{K},
$$

we obtain the stated expression. 
\end{proof}

\subsection{Proof of Corollary \ref{cor:corollary-up-size-continuous}}
\label{ap:corollary-up-size-continuous}

\begin{proof}[Proof of Corollary \ref{cor:corollary-up-size-continuous}]
Recall that the cardinality of the top-$K$ Rashomon set is bounded by

$$
|\mathcal R_\epsilon^K|
\leq
\sum_{m=0}^{M}
\binom{K}{m}\binom{N-K}{m},
$$

where $M=\min\{\lfloor\epsilon/\gamma_K\rfloor,K\}$. Recall that $m$ denotes the number of swaps between items inside and outside the optimal top-$K$ set. Since the optimal set contains $K$ items and its complement contains $N-K$ items, at most $\min(K,N-K)$ swaps are possible. Accordingly, the number of sets obtained through exactly $m$ swaps is $\binom{K}{m}\binom{N-K}{m}$, which is zero whenever $m>K$ or $m>N-K$.

Consequently, the cardinality bound can be expressed as $|\mathcal R_\epsilon^K| \leq 1+ \sum_{m=1}^{\min(K,N-K)} \binom{K}{m}\binom{N-K}{m} \cdot \boldsymbol{1} \big \{ \gamma_K\leq\frac{\epsilon}{m} \big \}$, where the initial term $1$ accounts for the optimal top-$K$ set, which incurs zero regret. 

Taking expectations on both sides and using the fact that $\mathbb{E}[\mathbf{1}\{\gamma_K\leq x\}]=F_{\gamma_K}(x)$, we obtain
$$
\mathbb E\!\left[|\mathcal R_\epsilon^K|\right]
\leq
1+\sum_{m=1}^{\min(K,N-K)}
\binom Km\binom{N-K}{m}
F_{\gamma_K}\!\left(\frac{\epsilon}{m}\right).
$$
which establishes the claimed upper bound.
\end{proof}

\subsection{Proof of Theorem \ref{th:boundary-gap}}
\label{ap:proof_boundary_gap}

\begin{proof}[Proof of Theorem \ref{th:boundary-gap}]

% By the Gaussian specialization of Theorem~\ref{th:boundary-gap-continuous}, the expected boundary gap admits the exact integral representation
% \begin{equation}
%     \overline{\gamma}_K = \sigma N\binom{N-1}{K} \int_{-\infty}^{\infty} \Phi(z)^{N-K-1}\phi(z) \cdot \left( \int_z^\infty \bigl(1-\Phi(w)\bigr)^K,dw \right)dz,
% \end{equation}
% where $\phi$ denotes the standard normal density. Equivalently, by
By the linearity of expectation, we know that $\overline{\gamma}_K = \mathbb{E}[\gamma_K]
=\mathbb E[Y_{[K]}]-\mathbb E[Y_{[K+1]}].$

Since $Y_i=\mu+\sigma Z_i$, where $Z_i\sim\mathcal N(0,1)$ independently, and the transformation $z\mapsto\mu+\sigma z$ preserves the ordering, we have $Y_{[r]}=\mu+\sigma Z_{[r]}$. Consequently, $\mathbb E[Y_{[r]}] =\mu+\sigma\mathbb E[Z_{[r]}].$

The expected normal order statistics do not generally admit elementary closed-form expressions. However, we can apply Blom's approximation \citep{blom1958statistical}:
\begin{equation}
    \mathbb E[Z_{[r]}] \approx \Phi^{-1}\left( \frac{N-r+\frac{5}{8}}{N+\frac{1}{4}} \right),
\end{equation}
where $Z_{[r]}$ denotes the $r$-th largest standard normal order statistic. Setting $r=K$ and $r=K+1$, respectively, gives
\begin{equation*}
    \mathbb E[Y_{[K]}] \approx \mu+\sigma \Phi^{-1}\left( \frac{N-K+\frac{5}{8}}{N+\frac{1}{4}} \right)
\end{equation*}

\begin{equation*}
    \mathbb E[Y_{[K+1]}] \approx \mu+\sigma \Phi^{-1}\left( \frac{N-K-\frac{3}{8}}{N+\frac{1}{4}} \right).
\end{equation*}

Subtracting these approximations cancels the common mean $\mu$ and yields the stated result.
    
\end{proof}

\subsection{Proof from Corollary \ref{cor:rashomon-size-approx-gaussian}}
\label{ap:rashomon-size-approx-gaussian}

\begin{proof}[Proof from Corollary \ref{cor:rashomon-size-approx-gaussian}]
% By the deterministic upper bound on the size of the top-$K$ Rashomon set,
% \begin{equation}
%     |\mathcal R_\epsilon^K| \leq 1+\sum_{m=1}^{\min(K,N-K)} \binom Km\binom{N-K}{m} \mathbf{1}\left\{\gamma_K\leq\frac{\epsilon}{m}\right\}.
% \end{equation}

% Taking expectations on both sides and using
% $\mathbb E[\mathbf{1}\{\gamma_K\leq x\}]=F_{\gamma_K}(x)$ gives

In Corollary \ref{cor:corollary-up-size-continuous} we defined the following upper bound on the expectation of the Rashomon set size for utilities that are sampled from a continuous distribution:

\begin{equation*}
    \mathbb E\!\left[|\mathcal R_\epsilon^K|\right] \leq 1+\sum_{m=1}^{\min(K,N-K)} \binom Km\binom{N-K}{m} F_{\gamma_K}\left(\frac{\epsilon}{m}\right).
\end{equation*}

We now approximate the boundary-gap CDF under the Gaussian assumption.
Let $Z_i=(Y_i-\mu)/\sigma$, so that the $Z_i\sim\mathcal N(0,1)$. Since affine transformations with positive scale preserve the ordering, $\gamma_K = \sigma\bigl(Z_{[K]}-Z_{[K+1]}\bigr).$

Let $\rho_N=K/N$, and suppose that $\rho_N\to\rho\in(0,1)$ as $N\to\infty$. The asymptotic theory of adjacent order-statistic spacings (Theorem 1 by \citet{nagaraja2015spacings}) implies
\begin{equation*}
    N f_Y(q_N)\gamma_K\xrightarrow{d}\operatorname{Exp}(1), \qquad q_N=F_Y^{-1}(1-\rho_N),
\end{equation*}
where $F_Y$ and $f_Y$ are the Gaussian CDF and density, respectively. The first expression implies that, when the number of utilities $N$ becomes large, the gap $\gamma_K$ between the $K$-th and $(K+1)$-th largest utilities behaves approximately like an exponentially distributed random variable after an appropriate rescaling. Besides, $F_Y^{-1}$ is the quantile function of the utlity distribution. Thus, the second expression implies that since the utilities are ordered in descending order, the $K$-th largest observation lies approximately at the quantile $1-K/N$ of the original distribution.

Moreover, if $X\sim\operatorname{Exp}(1)$, then $\Pr(X\leq x)=1-e^{-x}, x\geq0.$ Consequently, for large $N$, we approximate
\begin{equation*}
\begin{aligned}
\Pr(\gamma_K\leq x)
&=\Pr\left(Nf_Y(q_N)\gamma_K
\leq Nf_Y(q_N)x\right)\\
&\approx 1-\exp\left(-Nf_Y(q_N)x\right).
\end{aligned}
\end{equation*}

That is, the boundary-gap CDF is approximated by $F_{\gamma_K}(x) \approx 1-\exp\bigl(-Nf_Y(q_N)x\bigr).$
Moreover, for Gaussian utilities, $q_N=\mu+\sigma\Phi^{-1}(1-\rho_N)$
and therefore $f_Y(q_N) =\frac{1}{\sigma} \phi\left(\Phi^{-1}(1-\rho_N)\right).$
Substituting $x=\epsilon/m$ yields
\begin{equation*}
    F_{\gamma_K}\left(\frac{\epsilon}{m}\right) \approx 1-\exp\left[ -\frac{N\epsilon}{m\sigma} \phi\left(\Phi^{-1}\left(1-\frac{K}{N}\right)\right) \right].
\end{equation*}

Finally, substituting this approximation into the expected-size bound
gives
\[
\begin{aligned}
\mathbb E\!\left[|\mathcal R_\epsilon^K|\right]
\ \lesssim\ 1
&+\sum_{m=1}^{\min(K,N-K)}
\binom Km\binom{N-K}{m}\\
&\quad\times
\left[
1-\exp\left(
-\frac{N\epsilon}{m\sigma}
\phi\left(\Phi^{-1}\left(1-\frac{K}{N}\right)\right)
\right)
\right].
\end{aligned}
\]
\end{proof}

\section{Proofs from Section \ref{sec:fairness-analysis}}

\subsection{Proof of Theorem \ref{th:multi-group-fairness-optim}}
\label{ap:proof-fairness-optim}

To prove Theorem \ref{th:multi-group-fairness-optim}, we first present a supporting lemma and then provide the formal proof of the theorem.

\begin{lemma}[Equivalence of Absolute Value Minimal Minimization and Slack Linearization \cite{bertsimas1997introduction, vandenberghe2004convex}]
\label{lem:slack}
For any real-valued expression $\Delta \in \mathbb{R}$, minimizing its absolute value $|\Delta|$ subject to non-negative costs is equivalent to introducing non-negative slack variables $d^+, d^- \ge 0$ such that $\Delta = d^+ - d^-$ and minimizing the sum $(d^+ + d^-)$.
\end{lemma}

\begin{proof}[Proof of Lemma~\ref{lem:slack}]
Any real number $\Delta \in \mathbb{R}$ can be uniquely decomposed into its positive and negative components: $\Delta = \Delta^+ - \Delta^-$, where $\Delta^+ = \max(0, \Delta) \ge 0$ and $\Delta^- = \max(0, -\Delta) \ge 0$. Under this decomposition, $|\Delta| = \Delta^+ + \Delta^-$. 

Replacing $\Delta^+$ and $\Delta^-$ with general non-negative decision variables $d^+$ and $d^-$ yields the linear equality $\Delta = d^+ - d^-$. Under an objective function that penalizes $(d^+ + d^-)$ with positive coefficients, any optimal solution strictly enforces complementary slackness (i.e., at most one of $d^+$ or $d^-$ can be strictly positive at optimality, making $d^+ d^- = 0$). Hence, at optimality, $d^+ = \max(0, \Delta)$ and $d^- = \max(0, -\Delta)$, guaranteeing that $d^+ + d^- = |\Delta|$.
\end{proof}

% \begin{theorem}[Derivation of the Binary Integer Linear Program]
% The conceptual optimization problem of finding a candidate subset $\mathcal{C}_K \subseteq \mathcal{C}$ within the $\epsilon$-level Rashomon set $\mathcal{R}_\epsilon^K$ that minimizes $L_1$-norm aggregate representation disparity across $A$ groups is exactly equivalent to the Binary Integer Linear Program \eqref{obj:multi_fairness}--\eqref{cons:slack_nonneg}.
% \end{theorem}

\begin{proof}[Proof of Theorem \ref{th:multi-group-fairness-optim}]
We derive the formulation in four steps:

\paragraph{Step 1: Set Representation via Binary Decision Variables}
Let % $\mathcal{C} = \{c_1, c_2, \dots, c_N\}$ 
$\mathcal{C} = \{1, 2, \dots, N\}$ 
be the pool of $N$ candidates. Any selection of a candidate subset $\mathcal{C}_K \subseteq \mathcal{C}$ can be represented as an $N$-dimensional binary vector $\boldsymbol{\alpha} = [\alpha_1, \alpha_2, \dots, \alpha_N]^T \in \{0, 1\}^N$, where:
\begin{equation}
    \alpha_i = \begin{cases} 1 & \text{if } i \in \mathcal{C}_K, \\ 0 & \text{if } i \notin \mathcal{C}_K. \end{cases}
\end{equation}
% \begin{equation}
%     \alpha_i = \begin{cases} 1 & \text{if } c_i \in \mathcal{C}_K, \\ 0 & \text{if } c_i \notin \mathcal{C}_K. \end{cases}
% \end{equation}
The fixed cardinality condition $|\mathcal{C}_K| = K$ directly maps to the linear equality constraint:
\begin{equation}
    \sum_{i=1}^N \alpha_i = K, \quad \alpha_i \in \{0, 1\} \quad \forall i \in \{1, \dots, N\},
\end{equation}
which establishes constraints \eqref{cons:top_k} and \eqref{cons:binary}.

\paragraph{Step 2: Linear Formulation of the Rashomon Utility Constraint}
Recall that the $\epsilon$-level Rashomon set $\mathcal{R}_\epsilon^K$ requires the utility loss of $\mathcal{C}_K$ relative to the Bayes optimal set $\setref$ to not exceed $\epsilon$:
\begin{equation}
    \sum_{y \in \setref \setminus \mathcal{C}_K} y - \sum_{y' \in \mathcal{C}_K \setminus \setref} y' \le \epsilon.
    \label{eq:raw_rashomon}
\end{equation}
Using the decision vector $\boldsymbol{\alpha}$, we partition the candidate pool into candidates initially selected in the optimal set ($\setref$) and those not selected ($\mathcal{C} \setminus \setref$):
\begin{enumerate}
    \item For $i \in \setref$, candidate $i$ is excluded from $\mathcal{C}_K$ if $\alpha_i = 0$, which is indicated by $(1 - \alpha_i) = 1$. The utility lost from dropped optimal candidates is $\sum_{i \in \setref} y_i (1 - \alpha_i)$.
    \item For $j \notin \setref$, candidate $j$ is included in $\mathcal{C}_K$ if $\alpha_j = 1$. The utility gained from newly added non-optimal candidates is $\sum_{j \notin \setref} y_j \alpha_j$.
\end{enumerate}
Substituting these expressions into \eqref{eq:raw_rashomon} yields the linear inequality constraint:
\begin{equation}
    \sum_{i \in \setref} y_i (1 - \alpha_i) - \sum_{j \notin \setref} y_j \alpha_j \le \epsilon,
\end{equation}
which derives constraint \eqref{cons:rashomon}.

\paragraph{Step 3: Linearization of Group Representation Difference}
Let $a_{i, g} = \mathbb{I}(c_i \in \mathcal{G}_g)$ denote group membership. The number of selected candidates belonging to group $\mathcal{G}_g$ under decision vector $\boldsymbol{\alpha}$ is $|\mathcal{C}_K \cap \mathcal{G}_g| = \sum_{i=1}^N a_{i, g} \alpha_i$. 

Thus, the Representation Difference for group $\mathcal{G}_g$ is:
\begin{equation}
    \textrm{RD}_g(\boldsymbol{\alpha}, \mathcal{G}_g) = \frac{1}{K} \sum_{i=1}^N a_{i, g} \alpha_i - \frac{|\mathcal{G}_g|}{N}.
\end{equation}
Applying Lemma~\ref{lem:slack} to $\Delta_g = \textrm{RD}_g(\boldsymbol{\alpha}, \mathcal{G}_g)$, we introduce non-negative slack variables $d_g^+, d_g^- \ge 0$ representing positive over-representation and negative under-representation, respectively:
\begin{equation}
    \frac{1}{K} \sum_{i=1}^N a_{i, g} \alpha_i - \frac{|\mathcal{G}_g|}{N} = d_g^+ - d_g^-, \quad \forall g \in \{1, \dots, A\},
\end{equation}
which derives equality constraints \eqref{cons:disparity_slack} and non-negativity bounds \eqref{cons:slack_nonneg}.

\paragraph{Step 4: Construction of the Objective Function}
The $L_1$-norm aggregate disparity across all $A$ groups measures total absolute deviation:
\begin{equation}
    f_{L_1}(\boldsymbol{\alpha}) = \|\textrm{\textbf{RD}}(\boldsymbol{\alpha})\|_1 = \sum_{g=1}^A \left| \textrm{RD}_g(\boldsymbol{\alpha}, \mathcal{G}_g) \right|.
\end{equation}
By Lemma~\ref{lem:slack}, since $d_g^+, d_g^- \ge 0$ are constrained such that $d_g^+ - d_g^- = \textrm{RD}_g(\boldsymbol{\alpha}, \mathcal{G}_g)$, minimizing $\sum_{g=1}^A |\textrm{RD}_g(\boldsymbol{\alpha}, \mathcal{G}_g)|$ is mathematically identical to minimizing the sum of the slack variables:
\begin{equation}
    \min_{\boldsymbol{\alpha}, \boldsymbol{d}^+, \boldsymbol{d}^-} \quad \sum_{g=1}^A \left( d_g^+ + d_g^- \right).
\end{equation}
This completes the derivation of objective function \eqref{obj:multi_fairness}.
\end{proof}

\subsection{Less Discriminatory Rankings Under the $L_\infty$-Norm}
% \subsection{Proof of Theorem \ref{th:multi-group-fairness-optim-linf}}
\label{ap:multi-group-fairness-optim-linf}

\begin{theorem}[Multi-Group Fair Top-$K$ Optimization in $\mathcal{R}_\epsilon^K$ under $L_\infty$-Norm]
Let $\alpha_i \in \{0, 1\}$ be a decision variable indicating whether candidate $c_i \in \mathcal{C}$ is selected into top-$K$ ($\alpha_i = 1$). Let $a_{i, g} = \mathbb{I}(c_i \in \mathcal{G}_g)$ be the group membership indicator. Finding the globally fairest ranking within the $\epsilon$-level Rashomon set $\mathcal{R}_\epsilon^K$ under $L_\infty$-norm disparity is equivalent to solving the Binary Integer Linear Program:
\begin{align}
    \min_{\boldsymbol{\alpha}, M} \quad & M \label{obj:multi_fairness_linf} \\
    \text{s.t.} \quad & \sum_{i=1}^N \alpha_i = K, \label{cons:top_k_inf} \\
    & \sum_{i \in \setref} y_i (1 - \alpha_i) - \sum_{j \notin \setref} y_j \alpha_j \le \epsilon, \label{cons:rashomon_inf} \\
    & \frac{1}{K} \sum_{i=1}^N a_{i, g} \alpha_i - \frac{|\mathcal{G}_g|}{N} \le M, \quad \forall g \in \{1, \dots, A\}, \label{cons:disparity_linf_pos_inf} \\
    & \frac{|\mathcal{G}_g|}{N} - \frac{1}{K} \sum_{i=1}^N a_{i, g} \alpha_i \le M, \quad \forall g \in \{1, \dots, A\}, \label{cons:disparity_linf_neg_inf} \\
    & \alpha_i \in \{0, 1\}, \quad \forall i \in \{1, \dots, N\}, \label{cons:binary} \\
    & M \ge 0. \label{cons:max_bound_nonneg_inf}
\end{align}
In this formulation, constraint \eqref{cons:top_k_inf} enforces cardinality, \eqref{cons:rashomon_inf} restricts outcomes to the $\epsilon$-level Rashomon set, and constraints \eqref{cons:disparity_linf_pos_inf} and \eqref{cons:disparity_linf_neg_inf} bound the absolute group representation disparity for each group by $M$.
\label{th:multi-group-fairness-optim-linf}
\end{theorem}

\begin{proof}[Proof of Theorem~\ref{th:multi-group-fairness-optim-linf}]
We prove the equivalence by establishing that the feasible solutions of the
binary integer linear program (BILP) correspond exactly to the rankings in
$\mathcal{R}_\epsilon^K$, and that the objective of the BILP represents the
$L_\infty$-norm group disparity.

For any binary selection vector
$\boldsymbol{\alpha}\in\{0,1\}^N$, define the group representation disparity
for group $g$ as
\begin{equation}
    \Delta_g(\boldsymbol{\alpha})
    =
    \frac{1}{K}\sum_{i=1}^N a_{i,g}\alpha_i
    -
    \frac{|\mathcal{G}_g|}{N}.
\end{equation}
The $L_\infty$-norm disparity is then given by
\begin{equation}
    \|\boldsymbol{\Delta}(\boldsymbol{\alpha})\|_\infty
    =
    \max_{g\in\{1,\ldots,A\}}
    |\Delta_g(\boldsymbol{\alpha})|.
    \label{eq:original_linf_problem}
\end{equation}

\paragraph{Feasibility and Rashomon-set membership.}
The binary constraints
$\alpha_i\in\{0,1\}$, together with
\begin{equation}
    \sum_{i=1}^N\alpha_i=K,
\end{equation}
ensure that exactly $K$ candidates are selected. Moreover, the Rashomon
constraint
\begin{equation}
    \sum_{i\in\setref}y_i(1-\alpha_i)
    -
    \sum_{j\notin\setref}y_j\alpha_j
    \leq\epsilon
\end{equation}
restricts the selected candidates to those whose loss difference relative
to the reference top-$K$ selection does not exceed $\epsilon$.
By the definition of $\mathcal{R}_\epsilon^K$, these constraints characterize
its feasible selection vectors. Thus, the BILP's feasible binary vectors
correspond exactly to the rankings in $\mathcal{R}_\epsilon^K$.

\paragraph{Linearization of the $L_\infty$-norm objective.}
Consider the original fairness optimization problem
\begin{equation}
    \min_{\boldsymbol{\alpha}\in\mathcal{R}_\epsilon^K}
    \|\boldsymbol{\Delta}(\boldsymbol{\alpha})\|_\infty.
    \label{eq:original_fairness_optimization}
\end{equation}
Introduce an auxiliary variable $M\geq 0$ to upper-bound the maximum
absolute group disparity. For any feasible selection vector
$\boldsymbol{\alpha}$, the condition
\begin{equation}
    |\Delta_g(\boldsymbol{\alpha})|\leq M,
    \qquad \forall g\in\{1,\ldots,A\},
\end{equation}
is equivalent to the pair of inequalities
\begin{equation}
    -M\leq\Delta_g(\boldsymbol{\alpha})\leq M,
    \qquad \forall g\in\{1,\ldots,A\}.
\end{equation}
Substituting the definition of $\Delta_g(\boldsymbol{\alpha})$ yields
\begin{align}
    \frac{1}{K}\sum_{i=1}^N a_{i,g}\alpha_i
    -
    \frac{|\mathcal{G}_g|}{N}
    &\leq M,
    \quad \forall g\in\{1,\ldots,A\},
    \label{eq:proof_pos}
    \\
    \frac{|\mathcal{G}_g|}{N}
    -
    \frac{1}{K}\sum_{i=1}^N a_{i,g}\alpha_i
    &\leq M,
    \quad \forall g\in\{1,\ldots,A\}.
    \label{eq:proof_neg}
\end{align}
These inequalities are precisely constraints
\eqref{cons:disparity_linf_pos_inf} and
\eqref{cons:disparity_linf_neg_inf}, respectively.

Consequently, for any fixed feasible $\boldsymbol{\alpha}$, the smallest
feasible value of $M$ is
\begin{equation}
    M(\boldsymbol{\alpha})
    =
    \max_{g\in\{1,\ldots,A\}}
    |\Delta_g(\boldsymbol{\alpha})|
    =
    \|\boldsymbol{\Delta}(\boldsymbol{\alpha})\|_\infty.
\end{equation}
Therefore, minimizing $M$ over the BILP's feasible region is equivalent
to minimizing the $L_\infty$-norm disparity over
$\mathcal{R}_\epsilon^K$.

Combining the feasibility characterization with the epigraph reformulation,
we conclude that any optimal solution
$(\boldsymbol{\alpha}^*,M^*)$ of the BILP satisfies
\begin{equation}
    \boldsymbol{\alpha}^*
    \in\mathcal{R}_\epsilon^K,
    \qquad
    M^*
    =
    \|\boldsymbol{\Delta}(\boldsymbol{\alpha}^*)\|_\infty,
\end{equation}
and $\boldsymbol{\alpha}^*$ minimizes the $L_\infty$-norm group disparity
over $\mathcal{R}_\epsilon^K$.
This establishes the claimed equivalence.
\end{proof}

The proposed formulation shares strong structural ties with several foundational problems in operations research and decision theory. Most prominently, the $L_\infty$-norm variant exemplifies a classic minimax (or Chebyshev) approximation problem, where the objective minimizes the worst-case disparity across discrete sub-populations. Additionally, the structure mirrors the multi-choice 0/1 Knapsack Problem under exact capacity bounds, wherein candidate items carry multiple binary category labels and the objective penalizes deviations from proportional group target representation $\frac{|\mathcal{G}_g|}{N}$. Finally, the formulation can be interpreted through the lens of goal programming, where group-level representation targets serve as soft goals, and the optimization seeks an equitable allocation that minimizes maximum goal deviation within the feasible solution space dictated by the Rashomon set boundary.

\section{Empirical Evaluation}
\label{ap:empirical-evaluation-extended}

\subsection{Datasets} \label{ap:datasets}

\begin{table}[]
\centering
\caption{List of the datasets employed in the empirical analysis from Section~\ref{sec:experiments}.}
\label{tab:datasets}
\begin{tabular}{@{}llllcc@{}}
\toprule
Name            & Num. of instances & Score attr.  & Sensitive attr.    & Retrieved form & Used in \\ \midrule
NASA Astronauts & 357               & Flight hours & Gender            &  \citep{sourceAstronauts}              &   \citep{stoyanovich2018online}      \\
COMPAS          & 14854             & Risk score   & Gender, Ethnicity &  \citep{sourceCompas}              &   \cite{detoni,behzad2025reconciling}       \\
HMDA            & 211507            & Loan amout   & Gender, Ethnicity &  \citep{sourceHDMA}              &   \citep{so2022beyond,black2024d}      \\
LawSAT          & 21792             & LSAT score   & Gender, Ethnicity &   \citep{sourceLawSAT}            &   \citep{zehlike2020matching}      \\
SAT             & 1000              & SAT score    & Gender            & \cite{sourceSAT}                &  \cite{zehlike2017fa,konig2026performative}       \\ \bottomrule
\end{tabular}
\end{table}

This section provides a detailed description of the datasets employed in the empirical analysis presented in Section~\ref{sec:experiments}. Table~\ref{tab:datasets} lists these datasets, including their size, score and sensitive attributes, sources, and a few related works that employed them. The datasets are the following:

\paragraph{NASA Astronauts} It comprises the complete list of NASA astronauts from 1959 up until this day. For each astronaut, the dataset includes information such as the gender, birth place and date, alma mater, etc. From these attributes, we only employ the gender and flight hours of the astronauts. 

\paragraph{COMPAS} 
COMPAS is a software used in some parts of the US to assess the likelihood of a defendant becoming a recidivist.
The dataset was generated in the ProPublica analysis on the bias of ML algorithms in criminal sentencing \citep{sourceCompas}. It contains 6172 observations and 14 features. 
For our experimentation, we averaged the risk scores of the observations corresponding to the same individual, leading to a total of 14.8K entries. Note that we employ the raw recidivism risk scores instead of the usual categorical values for this attribute.  

\paragraph{HMDA} The dataset was sourced from the Home Mortgage Disclosure Act (HMDA) by the FFIEC. 
Originally enacted by the US Congress in 1975, the HMDA requires financial institutions to maintain, report, and publicly disclose loan-level information about mortgages. In our experimentation, we selected the mortgage data corresponding to the state of New York in the year 2025 (the newest data available at the time of submitting this paper).  From this data, we selected the loan amount as the score attribute and gender and ethnicity as the sensitive ones.

\paragraph{Law SAT} The Law School Admission Council conducted a survey across 163 law schools in the US in 1998 \cite{kusner2017counterfactual}. The dataset contains information on more than 21K law students such as their entrance exam scores (LSAT score), grade-point averages collected prior to law school, and their first year average grade. We employ the LSAT score and use the gender and ethnicity as sensitive attributes.

\paragraph{SAT}  The Educational Testing Service (ETS) collected data from 1,000 students attending a US college \cite{sourceSAT}. The dataset contains information about students' academic performance, including their verbal and mathematics SAT percentiles, high-school grade-point average (GPA), and first-year college GPA. We use the SAT scores and gender as sensitive attribute.

\subsection{Hardware Details} \label{ap:hardware}

All experiments were run in the same TUXEDO InfinityBook Pro (Gen 7, MK
1) laptop with a 12th Gen Intel Core i7-12700H CPU and 16GB of RAM.

\subsection{Inclusion Instability} \label{ap:instability}

\begin{figure}
    \centering
    \includegraphics[width=0.9\linewidth]{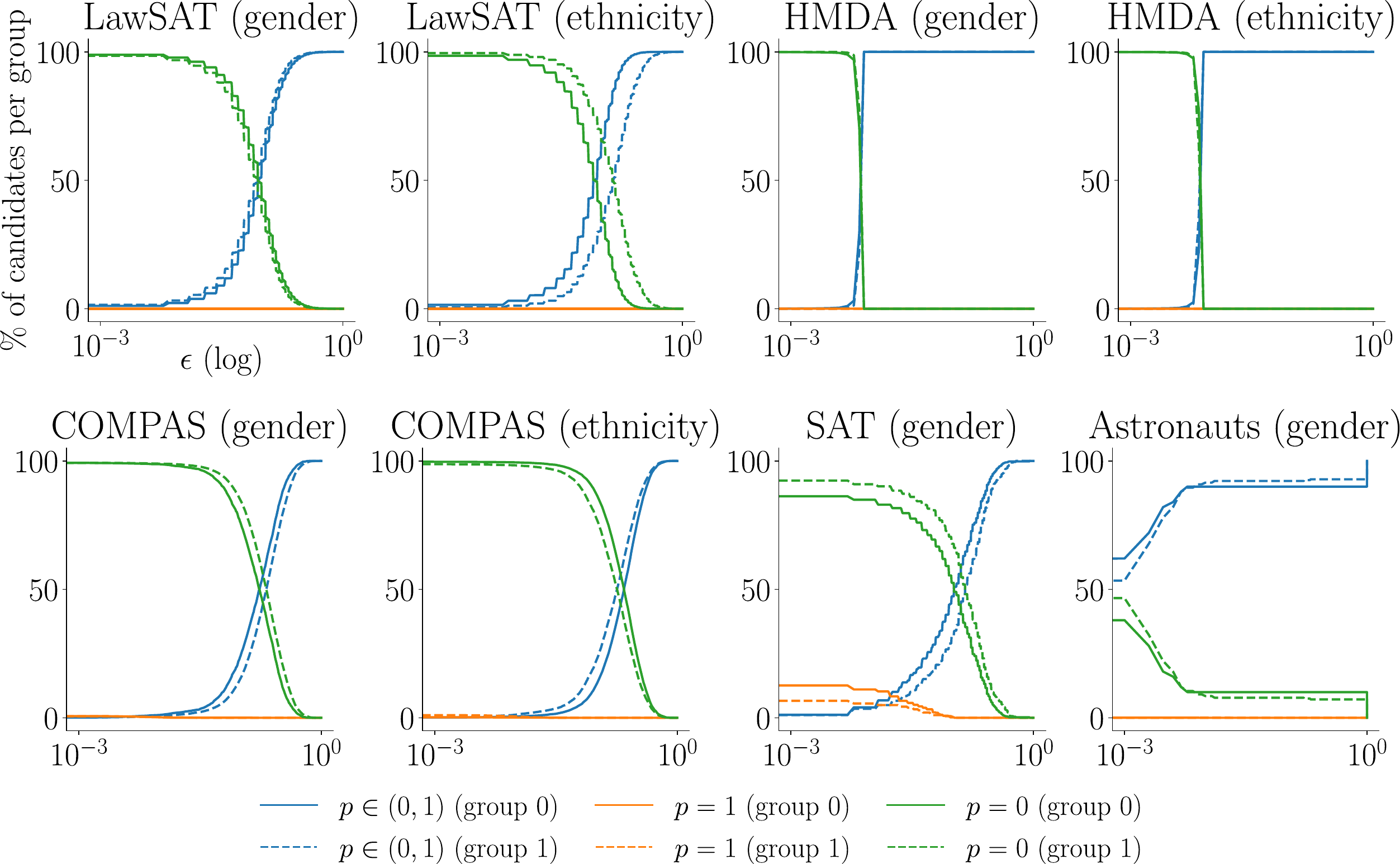}
    \caption{Percentage of individuals in each group for different inclusion probabilities in the top-K ($p_j^{(\epsilon)} = 0$, $p_j^{(\epsilon)} = 1$, and $p_j^{(\epsilon)} \in (0, 1)$) and different $\epsilon$ values. This figure includes the same analysis presented in Section~\ref{sec:experiments} for all of the considered datasets and sesible attributes. The size of the top-K was set to $K=100$ for all plots.}
    \label{fig:apx-instability}
\end{figure}

\begin{figure}
     \centering
     \begin{subfigure}[b]{0.49\textwidth}
         \centering
         \includegraphics[width=\textwidth]{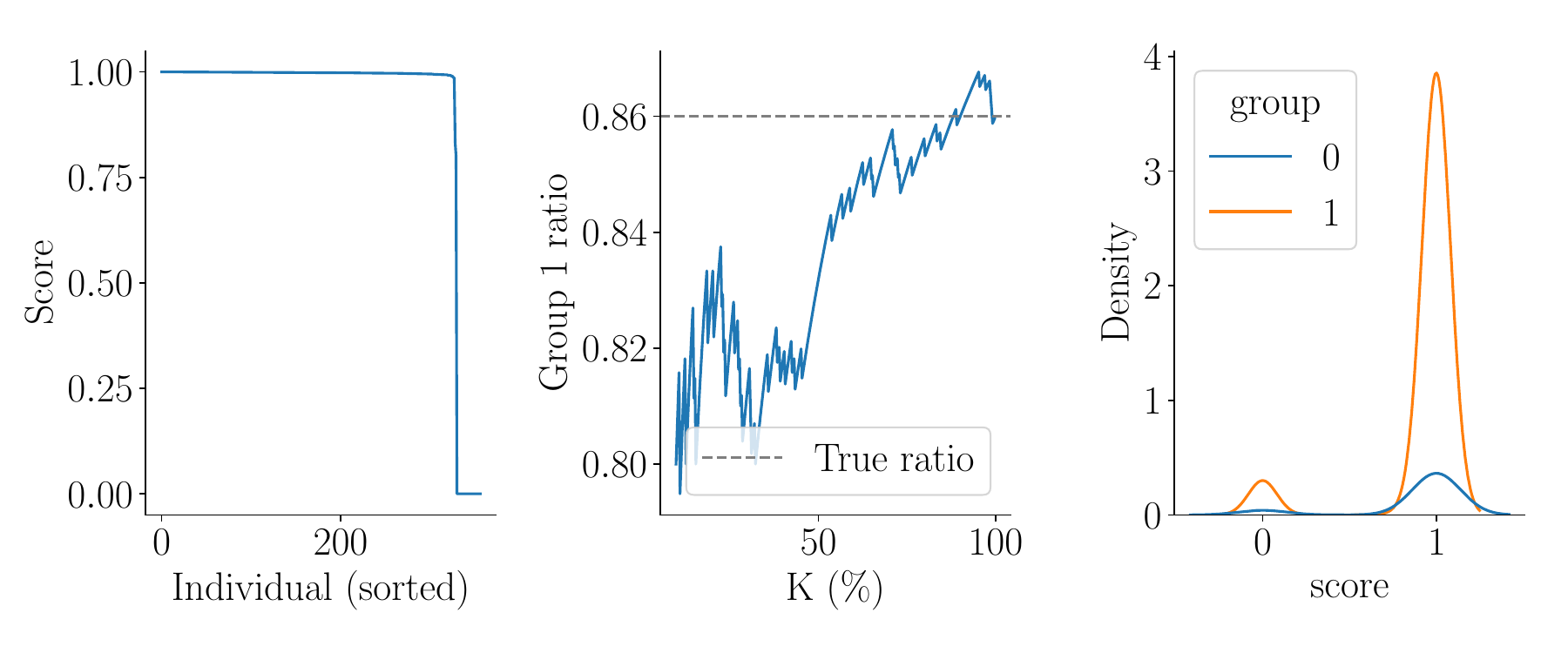}
         \caption{Astronauts (gender).}
         \label{fig:score-astro-gender}
     \end{subfigure}
     % \hfill
     \begin{subfigure}[b]{0.49\textwidth}
         \centering
         \includegraphics[width=\textwidth]{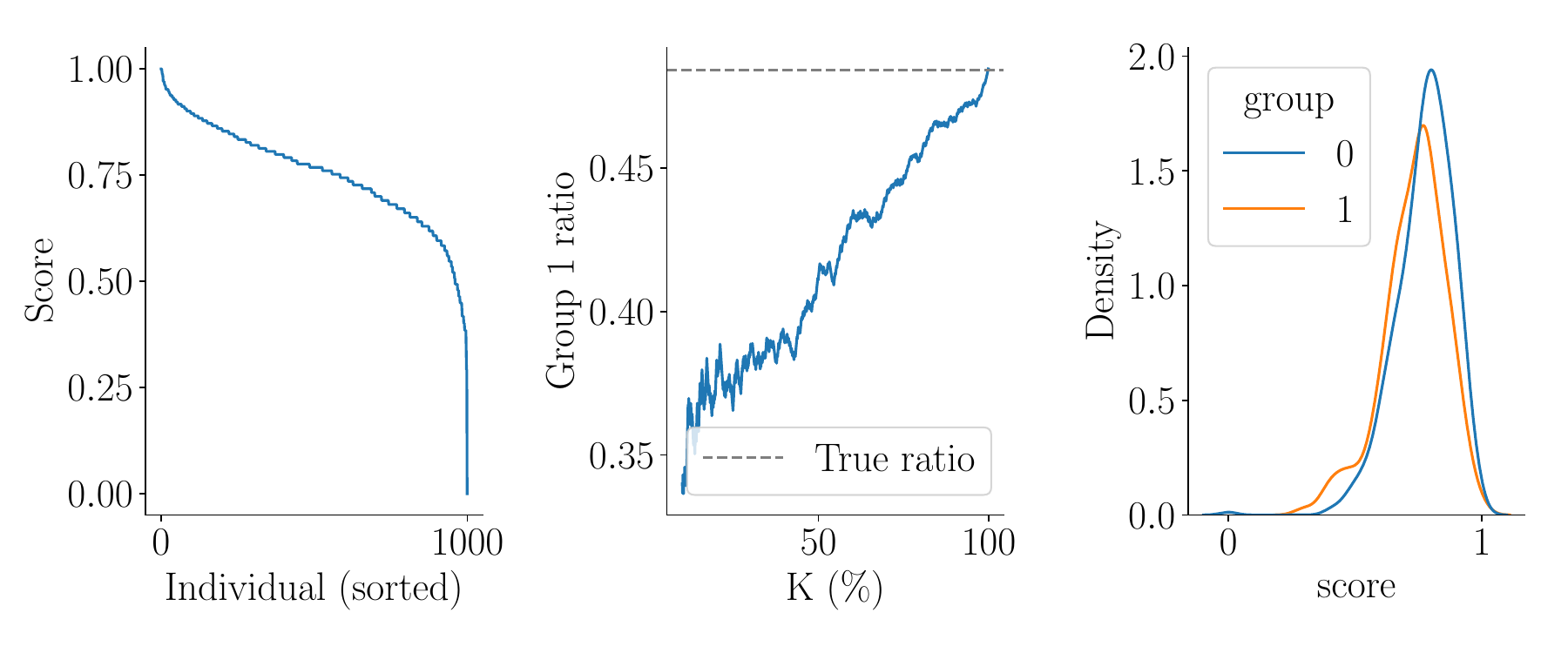}
         \caption{SAT (gender).}
         \label{fig:score-sat-gender}
     \end{subfigure}
     \\
     \begin{subfigure}[b]{0.49\textwidth}
         \centering
         \includegraphics[width=\textwidth]{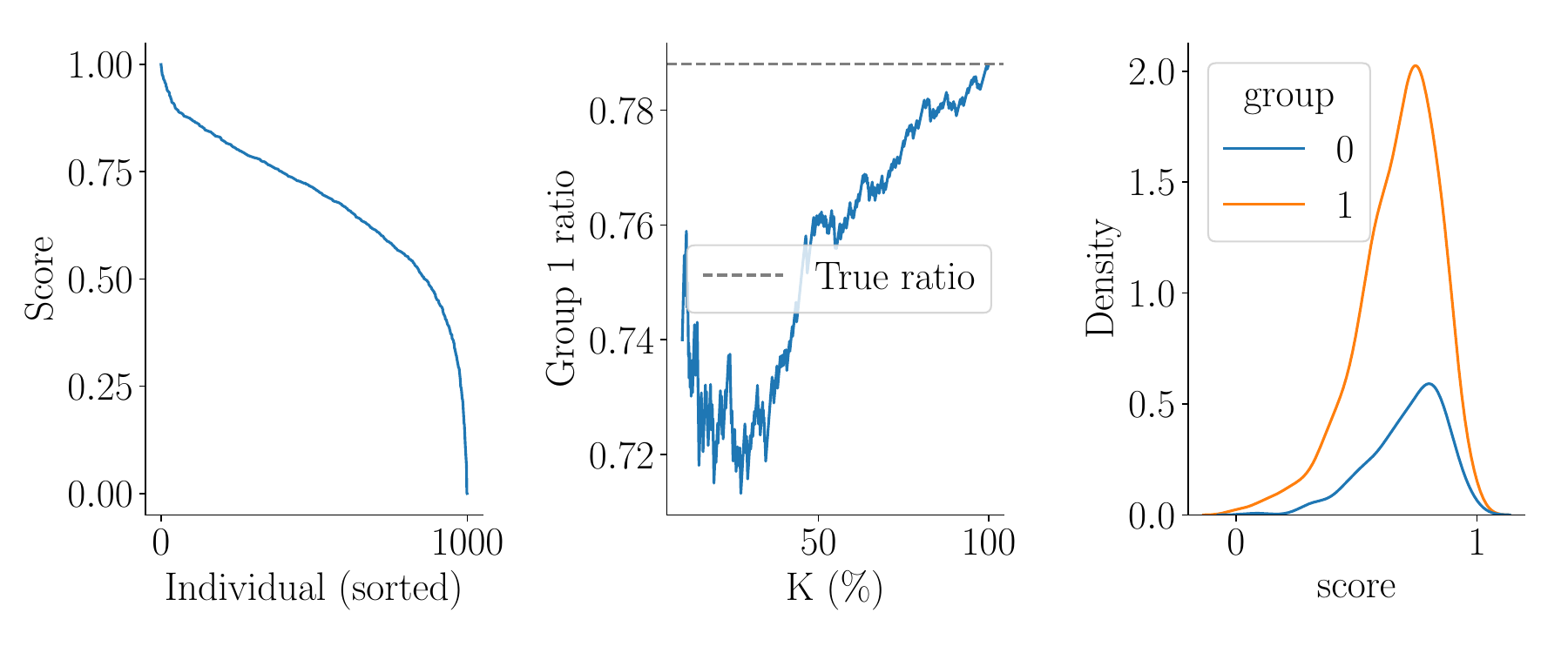}
         \caption{COMPAS (gender).}
         \label{fig:score-compas-gender}
     \end{subfigure}
     %  \hfill
     \begin{subfigure}[b]{0.49\textwidth}
         \centering
         \includegraphics[width=\textwidth]{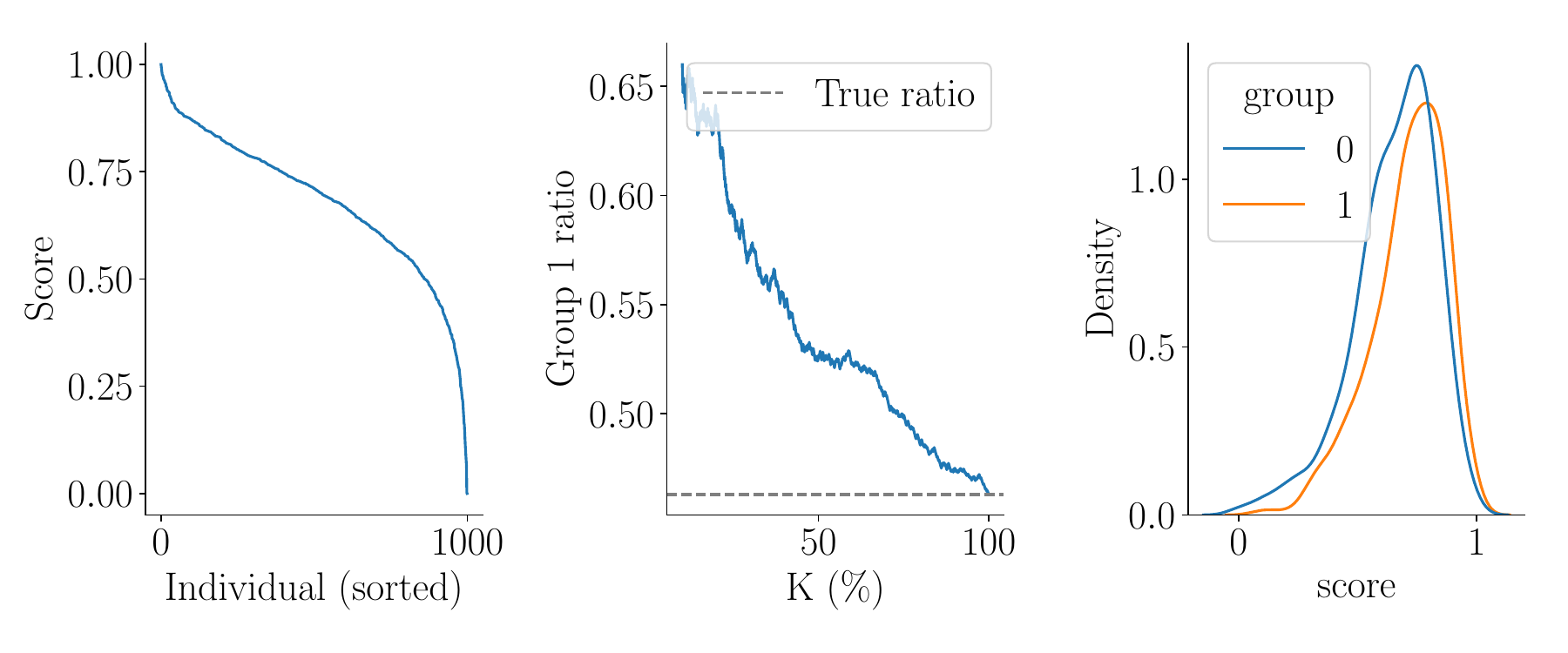}
         \caption{COMPAS (ethnicity).}
         \label{fig:scores-compas-ethnicity}
     \end{subfigure}
     \\
     \begin{subfigure}[b]{0.49\textwidth}
         \centering
         \includegraphics[width=\textwidth]{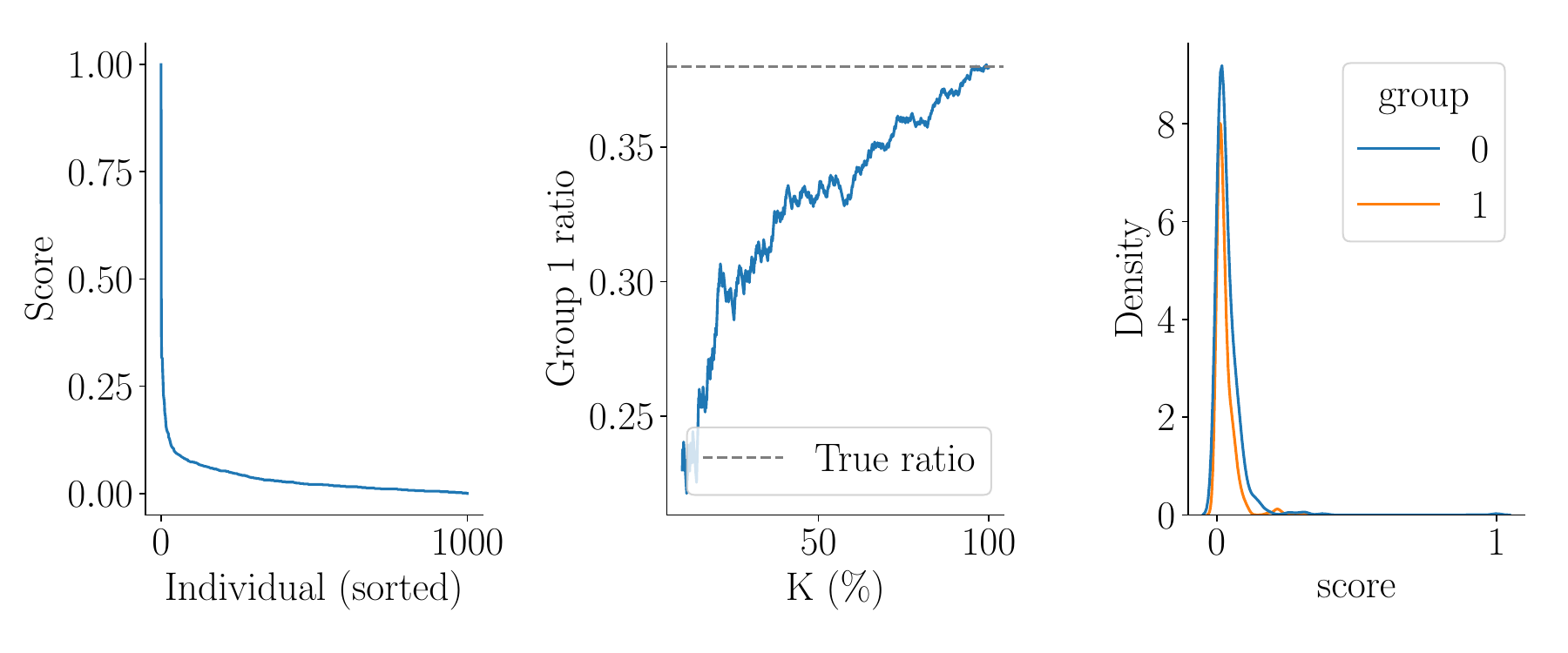}
         \caption{HMDA (gender).}
         \label{fig:scores-hdma-gender}
     \end{subfigure}
     % \hfill
     \begin{subfigure}[b]{0.49\textwidth}
         \centering
         \includegraphics[width=\textwidth]{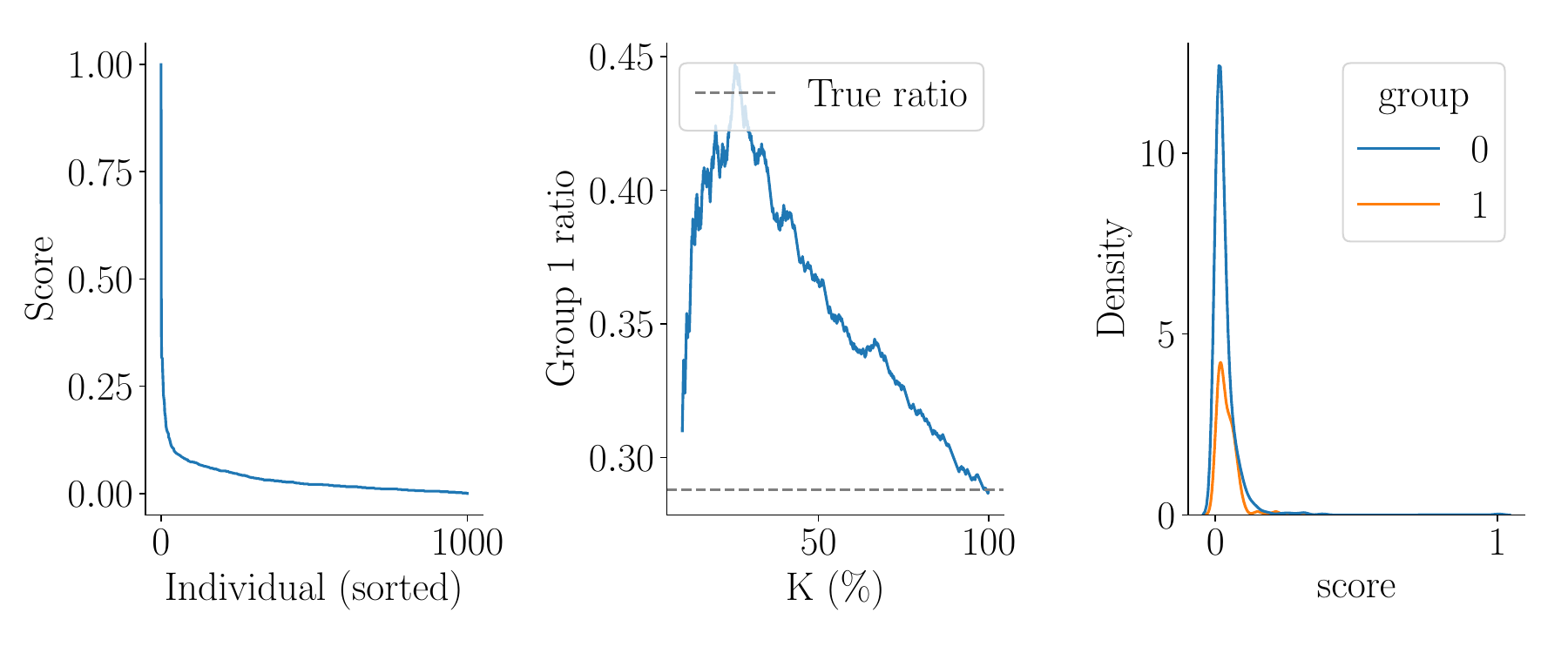}
         \caption{HMDA (ethnicity).}
         \label{fig:scores-hdma-ethnicity}
     \end{subfigure}
     \\
     \begin{subfigure}[b]{0.49\textwidth}
         \centering
         \includegraphics[width=\textwidth]{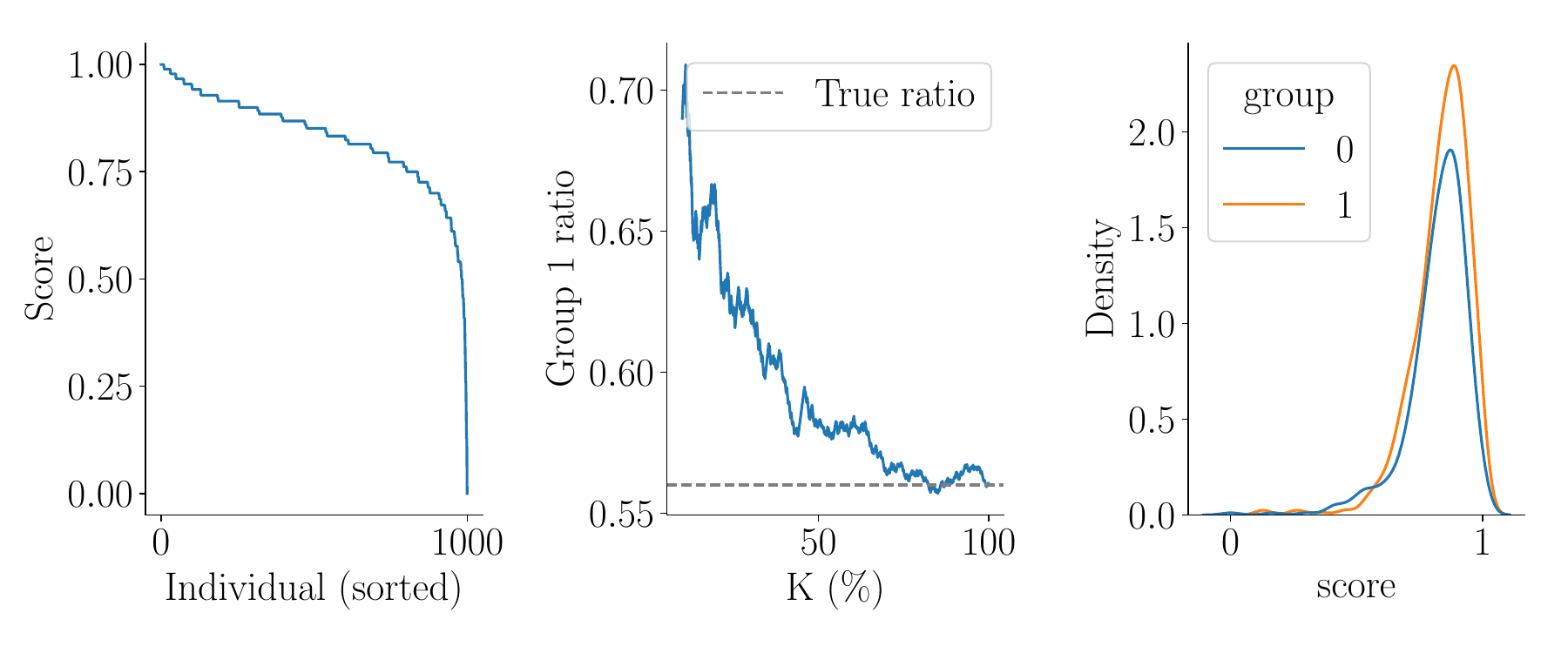}
         \caption{Law SAT (gender).}
         \label{fig:scores-lsat-gender}
     \end{subfigure}
     % \hfill
     \begin{subfigure}[b]{0.49\textwidth}
         \centering
         \includegraphics[width=\textwidth]{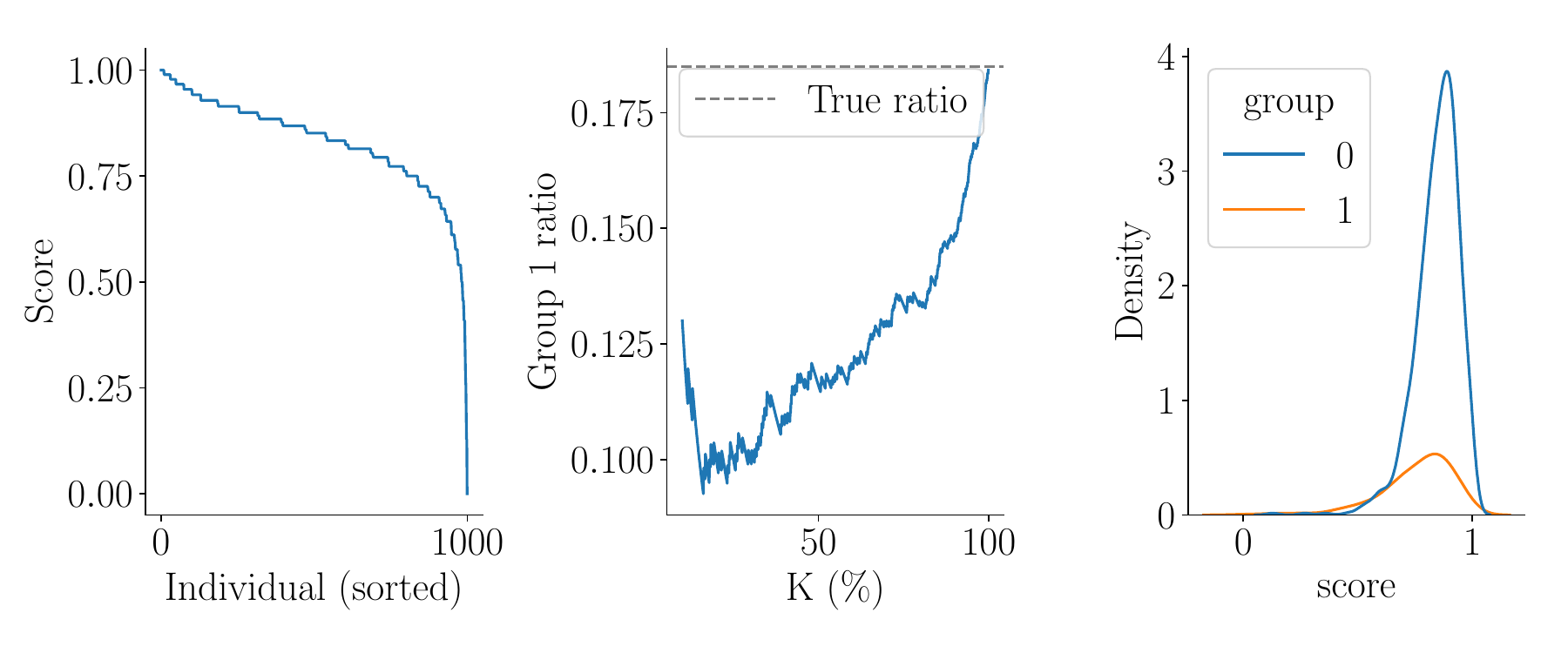}
         \caption{Law SAT (ethnicity).}
         \label{fig:scores-lsat-ethnicity}
     \end{subfigure}
     \caption{Each figure shows three plots, from left to right: the score of each individual, sorted from high to low; the ratio of individuals from group 1 in the Bayes top-K (i.e., scores ordered from high to low according to the true scores) for different values of K (percentages with respect to the total population size); the estimated density of the scores for each group. All plots employ the data preprocessing and score normalization mentioned in Section~\ref{sec:experiments}.}
     \label{fig:apx-all-scores}
\end{figure}

Due to main-paper page limits, we deferred the inclusion probability plots omitted from Figure~\ref{fig:placeholder} to Figure~\ref{fig:apx-instability} in this appendix. For the \textit{gender} sensitive attribute in the \texttt{LawSAT} dataset, the gap between the inclusion percentages for $p_j^{(\epsilon)}=0$ and $p_j^{(\epsilon)} \in (0, 1)$ is smaller than for the sensitive attribute of \textit{ethnicity}, suggesting greater inclusion stability in the Rashomon set of top-$K$ sets across gender groups than for sensitive groups defined by the ethnicity. Conversely, for the \texttt{COMPAS} dataset, this gap remains considerably more consistent across the sensitive attributes evaluated. Finally, the inter-group gap is very narrow for \texttt{HMDA}, where the percentages for inclusion probabilities $p_j^{(\epsilon)}=0$ and $p_j^{(\epsilon)} \in (0, 1)$ take extreme values—$\approx 100\%$ and $\approx 0\%$, respectively—that flip sharply at $\epsilon=0.007$. We hypothesize that this behavior is driven by the relatively small value of $K$ relative to the dataset size (approximately $0.047\%$ of the dataset, see Table~\ref{tab:datasets}) alongside the extreme score distribution of the dataset (see Figures~\ref{fig:scores-hdma-gender} and~\ref{fig:scores-hdma-ethnicity}): few individuals enjoy large loans, while the rest have much lower ones.

\subsection{Score Distributions} \label{ap:score-distributions}

Figure~\ref{fig:apx-all-scores} provides the score data analysis from Figure~\ref{fig:dataset-ch-scores} in Section~\ref{sec:experiments} for the remaining combinations of datasets and sensitive attributes.

By observing the ordered scores (leftmost plot in all subfigures from Figure~\ref{fig:apx-all-scores}), we notice two general trends regarding the area under the score curve. All datasets have a large area under the curve except HDMA, suggesting a huge imbalance between a few individuals (the ones with the greatest scores) and the majority (with low scores). As discussed in Appendix~\ref{ap:instability}, this imbalance can lead to more severe instability issues, inducing an additional source of unfairness.  

When analyzing group 1 ratios in top-K sets of different sizes (middle plot in subfigures from Figure~\ref{fig:apx-all-scores}), we consistently observe a substantial gap between the true and (Bayes) top-K ratios. For instance, see the center plot in Figure~\ref{fig:scores-compas-ethnicity}; for initial values of $K$, the gap between the true and top-K ratios is larger than 15\%. Disparities are always the largest before reaching the top-K size corresponding to 50\% of the dataset. Besides, in most cases, disparities are generally highest for smaller values of $K$, suggesting that fairness concerns become more pronounced in highly competitive settings.

Finally, we observe that most of the score distributions (rightmost plot in subfigures from Figure~\ref{fig:apx-all-scores}) are unimodal,  with the exception of group 0 in the Astronauts dataset (see Figure~\ref{fig:score-astro-gender}). This observation suggests that the theoretical results derived
%techniques such as those applied 
in Section~\ref{sec:decision-boundary} (see        Corollary~\ref{coro:set-size-normal}) and Section~\ref{sec:rashomon-empirical-insights} to analyze the properties of top-$K$ Rashomon sets might be interesting and useful for characterizing real-world ranking problems.

\end{document}